\pdfoutput=1

\documentclass[11pt]{article}

\usepackage[final]{acl}

\usepackage{times}
\usepackage{latexsym}

\usepackage[T1]{fontenc}

\usepackage[utf8]{inputenc}

\usepackage{microtype}
\usepackage{booktabs}
\usepackage{inconsolata}

\usepackage{graphicx}
\usepackage{multirow}

\NewDocumentCommand{\lifu}{ mO{} }{\textcolor{red}{\textsuperscript{\textit{Lifu}}\textsf{\textbf{\small[#1]}}}}

\NewDocumentCommand{\jy}{ mO{} }{\textcolor{brown}{\textsuperscript{\textit{Jingyuan}}\textsf{\textbf{\small[#1]}}}}

\NewDocumentCommand{\zian}{ mO{} }{\textcolor{teal}{\textsuperscript{\textit{Zian}}\textsf{\textbf{\small[#1]}}}}

\NewDocumentCommand{\qin}{ mO{} }{\textcolor{orange}{\textsuperscript{\textit{Qin}}\textsf{\textbf{\small[#1]}}}}

\NewDocumentCommand{\xf}{ mO{} }{\textcolor{purple}{\textsuperscript{\textit{Xiaofei}}\textsf{\textbf{\small[#1]}}}}

\definecolor{lightblue}{rgb}{0.0, 0.5, 1.0}
\NewDocumentCommand{\minqian}{ mO{} }{\textcolor{lightblue}{\textsuperscript{\textit{Minqian}}\textsf{\textbf{\small[#1]}}}}

\newcommand*\system{\textsc{MetaScientist}}

\title{\system{}: A Human-AI Synergistic Framework for Automated Mechanical Metamaterial Design}
\author{
Jingyuan Qi$^{* 1}$, , Zian Jia$^{* 2,3}$, Minqian Liu$^1$, Wangzhi Zhan$^1$, Junkai Zhang$^4$ \\
\textbf{Xiaofei Wen$^5$, Jingru Gan$^4$, Jianpeng Chen$^1$, Qin Liu$^5$, Derek Ma$^4$} \\
\textbf{Bangzheng Li$^5$, Haohui Wang$^1$, Adithya Kulkarni$^1$} \\
\textbf{Muhao Chen$^5$, Dawei Zhou$^1$, Ling Li$^3$, Wei Wang$^4$, Lifu Huang$^5$} \\
$^1$Virginia Tech \hspace{10pt} $^2$Princeton University \hspace{10pt} $^3$University of Pennsylvania  \\
$^4$University of California, Los Angeles \hspace{10pt} $^5$University of California, Davis\\
\texttt{Demo link: \url{http://zhoulab-1.cs.vt.edu:5557/}} \\
\thanks{These authors contributed equally.}
}

\begin{document}
\maketitle

\begin{abstract}
The discovery of novel mechanical metamaterials, whose properties are dominated by their engineered structures rather than chemical composition, is a knowledge-intensive and resource-demanding process. 
To accelerate the design of novel metamaterials, we present \system{}, a human-in-the-loop system that integrates advanced AI capabilities with expert oversight with two primary phases
: (1) hypothesis generation, where the system performs complex reasoning to generate novel and scientifically sound hypotheses, supported with domain-specific foundation models and inductive biases retrieved from existing literature; 
(2) 3D structure synthesis, where a 3D structure 
is synthesized with a novel 3D diffusion model based on the textual hypothesis 
and refined it with a LLM-based refinement model to achieve better structure properties.
At each phase, domain experts iteratively validate the system outputs, and provide feedback and supplementary materials to ensure the alignment of the outputs with scientific principles and human preferences. 
Through extensive evaluation from human scientists, \system{} is able to deliver novel and valid mechanical metamaterial designs that have the potential to be highly impactful in the metamaterial field\footnote{A demo video for \system{} can be accessed at: \url{https://drive.google.com/file/d/19HYCRYgEeuShJZgBUdSGI-l8KD-cfBAp/view?usp=sharing}}.

\end{abstract}
\vspace{-1mm}
\section{Introduction}
\label{sec:introduction}
\vspace{-1mm}

Metamaterials are microstructured materials whose properties go beyond those of the ingredient materials, often possessing unusual mechanical and/or functional properties~\cite{kadic20193d,bertoldi2017flexible,jia2020engineering,jiao2021artificial,bauer2017nanolattices}.  The advancements in metamaterials have fostered a vast amount of technological innovations across various key domains, such as structural, sensing, actuation, and multifunctional materials innovations.
However, human scientists are facing two critical challenges in designing and discovering novel metamaterials.

First, as the field of metamaterial expands exponentially, it becomes increasingly challenging for researchers to keep pace with the rapid accumulation of knowledge. 
For instance, the number of metamaterial-related publications has grown from 156 papers in 2000, 7,170 papers in 2010, to over 30,000 papers in 2023 based on Google Scholar statistics. 
The rapid proliferation of research has created significant barriers to effectively 
understanding and building upon existing work, ultimately slowing down the pace of technological innovation. Second, existing workflows for metamaterial design impose substantial cognitive and resource burdens throughout their multi-stage development process. From initial conceptualization through theoretical modeling, computational simulation, and experimental validation, each phase demands significant human expertise and creativity while incurring considerable time and financial costs. These combined constraints severely limit both the exploration of new concepts and the scaling of successful designs for practical applications.

\begin{figure*}[htbp]
	\centering
	\includegraphics[width=0.98\linewidth, trim={0 0 0 0},clip ]{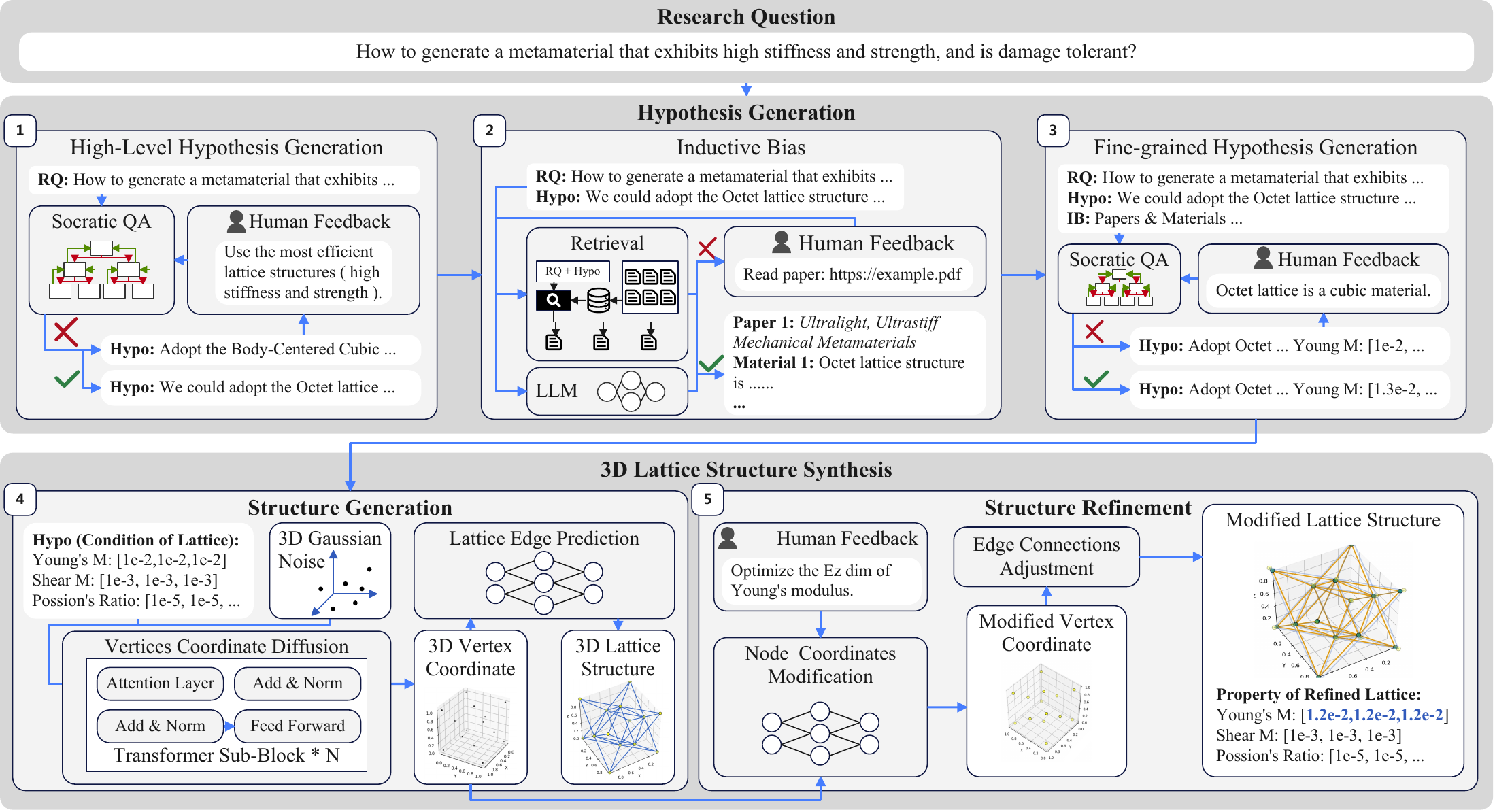}
    \vspace{-2mm}
	\caption{\textbf{Overview} of \system{}. }
    \vspace{-5mm}
\label{fig_overview}
\end{figure*}

Recent breakthroughs in large language models (LLMs) have inspired the development of \textit{autonomous scientists} to automate aspects of the research process, such as generating novel research ideas and outlining implementation procedures~\cite{zhang2024comprehensive}. However, in specialized scientific domains like metamaterial design, several recent studies~\cite{hong2023chatgpt,park2024can,deb2024chatgpt} have highlighted significant limitations even in state-of-the-art models like GPT-4o~\cite{openai2024gpt4ocard}. While these models can \textit{boldly} synthesize vast amounts of scientific knowledge and propose testable hypotheses, they are prone to high factual error rates, often stemming from confusion caused by similar nomenclatures across different fields or lack of rigorous logical reasoning and mathematical derivation capabilities, which are critical for scientific research. In addition, these models typically fail to assess the plausibility of their hypotheses effectively, and their proposals are often overly generic, offering limited practical value for real-world experiments.\footnote{We provide examples in Appendix~\ref{sec:eval_example} to illustrate the key limitations of GPT-4o for automated metamaterial design.}

To tackle these challenges, we introduce \system{}, a 
human-in-the-loop system for assisting and accelerating
novel metamaterial design, with a particular focus on mechanical metamaterials (e.g., lattice-based structures with superior mechanical properties). 
As depicted in Figure~\ref{fig_overview}, \system{} comprises two key phases: \textbf{(1) Hypothesis Generation}: Given an input research question, we perform 
complex reasoning upon evidence and inductive biases retrieved from a large collection of materials science literature to generate novel hypotheses.
\textbf{(2) 3D Structure Synthesis}: We build a 3D diffusion model to transform the generated textual hypotheses into 3D metamaterial structures, and then refine the generated structures 
with a LLM-based refinement model to achieve better structure properties (e.g., symmetry and periodicity)\footnote{Though Metamaterials can have diverse types of structure morphologies, we mainly focus on lattice metamaterials as they are common and have a large design space.}. 
Throughout both phases, we involve 
human scientists to actively interact with our system and iteratively provide diverse types of feedback, including cutting-edge insights, error correction, and supplementary materials, to 
ensure the generated hypotheses are feasible, novel, and aligned with 
up-to-date domain knowledge in metamaterials. 
Our contributions are summarized as follows:
\begin{itemize}
\vspace{-3mm}
    \item We introduce \system{}, the first complete hypothesis generation system to facilitate novel metamaterial design by automating knowledge and resource-intensive procedures.
\vspace{-7mm}
    \item \system{} involved domain experts in the loop to iteratively incorporate high-quality feedback from human scientists to ensure the validity and creativity of hypotheses.
\vspace{-3mm}
    \item With extensive evaluation and case studies, 
    we demonstrate \system{} has significant potential to advance AI-assisted metamaterial design and reduce the extensive cost and cognitive load required for human scientists.
\end{itemize}



\vspace{-5mm}
\section{Implementation of \system{}}
\label{sec:method}
\vspace{-1mm}

\subsection{Foundation Model for Metamaterials Science}
\label{sec:foundation_model}

While existing general-purpose LLMs excel at a wide range of NLP tasks, they often struggle in highly specialized domains such as materials science~\cite{song-etal-2023-honeybee}.
To address this challenge, we first collect 5,611 research papers and books from Google Scholar based on a list of keywords that are provided by a domain expert and related to metamaterial design or general domain knowledge of materials science. We then design a comprehensive data-cleaning process and convert them into plain text.\footnote{We include more details for data collection and statistics in Appendix~\ref{app:dataclean}} Based on this corpus, we finetune an 
open-sourced LLM, i.e., \texttt{Llama3-8B-Instruct}, using the Low-Rank Adaptation (LoRA) technique~\cite{hu2022lora}, and tailor the model to the metamaterials science domain. To qualitatively evaluate this model, we ask a domain expert to provide a list of question-answer pairs and compare it against an open-source LLM baseline.\footnote{Additional training and evaluation details for our domain-specific foundation model are provided in Appendix~\ref{app:modeltrain}.}

\subsection{Hypothesis Generation}

Generating novel, feasible, and scientifically sound hypotheses requires strong reasoning capability based on existing knowledge and literature. To this end, we equip our domain-specific foundation model in \S\ref{sec:foundation_model} 
with a recent state-of-the-art complex reasoning framework, i.e., Socratic Questioning~\cite{socraticqa}, to generate novel hypotheses. Specifically, the hypothesis generation process consists of three key steps.

\vspace{-2mm}
\paragraph{High-Level Hypothesis Generation}
Since the properties of lattice materials are dominated by their structure, in the first step, we focus on generating a high-level hypothesis that is novel and reasonable with normalized materials properties.
With the Socratic Questioning module, we first perform a \textbf{top-down exploration} process that recursively decomposes a complex research question, e.g., \textit{``how to generate a metamaterial that exhibits high stiffness and strength, and is damage tolerant''}, into several simpler sub-questions, such as \textit{``what properties should a material have to be damage tolerant''}, and \textit{``what are the common techniques to enhance both stiffness and strength''}. With the solutions of the sub-questions proposed by LLM itself, e.g., \textit{``a material should have high Young's modulus and Shear modulus to be damage tolerant''}, we employ a \textbf{bottom-up backtracking} process selectively leverage these answers to infer the hypothesis to solve the original research question, e.g., \textit{``Octet or Kelvin structures might be promising to solve the question''}.  

While the Socratic Questioning module can generate seemingly reasonable hypotheses, our preliminary experiments show that these automatically generated high-level hypotheses still suffer from several critical issues, including hallucinations, limited novelty, or deviating from desired directions. Therefore, we involve domain experts to provide feedback, either natural language feedback or relevant papers,
which are then incorporated as additional context when solving the original research question during the bottom-up backtracking process, encouraging the model to generate more valid and novel hypotheses that are aligned with human expectations.



\vspace{-2mm}
\paragraph{Inductive Bias Extraction}
The hypotheses proposed by human scientists are inherently shaped by inductive biases, i.e., existing knowledge and initial assumptions. A valid and novel hypothesis needs to excel in two aspects: it must be grounded in established knowledge while offering distinct contributions beyond existing literature. 
Thus, we propose an inductive bias extraction module that extracts relevant knowledge from multiple sources, including (1) \textit{internal inductive bias} which is elicited from LLM itself by prompting it based on an expert-written taxonomy\footnote{More details are provided in Appendix~\ref{app:hypo_gen}}, e.g., 
\textit{Based on the taxonomy, compile a list of known lattice structures relevant to the hypothesis...}; 
and (2) \textit{external inductive biases}, which are summaries of relevant articles retrieved from our collected literature corpus introduced in \S\ref{sec:foundation_model}. In this process, we use the concatenation of an original research question and a high-level hypothesis as the query,  \texttt{ColBERT}~\cite{ColBERT} as the text encoder, and L2 as the distance metric to retrieve top-\textit{K} relevant articles to the query. We then use GPT-4o to locate the relevant content to the query within each paper and summarize it into one paragraph for each article.

The inductive biases may contain information that is outdated or irrelevant to the query, thus we involve domain experts to either directly refine the outputs by removing undesired inductive biases or provide feedback to improve the retrieval process, such as providing additional keywords for more accurate retrieval, directly uploading relevant papers, or providing textual supplementary materials.



\vspace{-2mm}
\paragraph{Fine-grained Hypothesis Generation}
We further leverage the inductive biases as additional evidence, i.e., feeding them together with the original research question as input to the Socratic Questioning module, to refine the initial hypothesis into a fine-grained description of features and attributes that inform the structure design of the new target metamaterial. 
In our system, we primarily consider three critical attributes, i.e., \textit{Young's Modulus}, \textit{Shear Modulus}, and \textit{Poisson's Ratio}, for later 3D structure generation.
As the model may not be able to capture all necessary lattice structure properties or the predicted attributes are not aligned with human experts' expectations, we allow human experts to enrich the dimensions of proposed lattice structure properties and adjust fine-grained details to meet specific requirements.
We present a complete and concrete case study in \S\ref{sec:case_study} to demonstrate how our \system{} system generates a fine-grained hypothesis step-by-step.

\subsection{3D Structure Synthesis}

Our next step is to generate a 3D lattice structure that aligns with the previously generated hypothesis. This phase consists of two steps: generating 3D structures conditioned on the lattice properties in the fine-grained hypothesis using a diffusion model, and further refining the generated structures based on human feedback with a refinement model.\footnote{Training details are provided in Appendix~\ref{app:3d_syn}.}

\begin{figure}[!tbp]
    \centering
    \includegraphics[width=0.95\linewidth, trim={0 0 0 0},clip ]{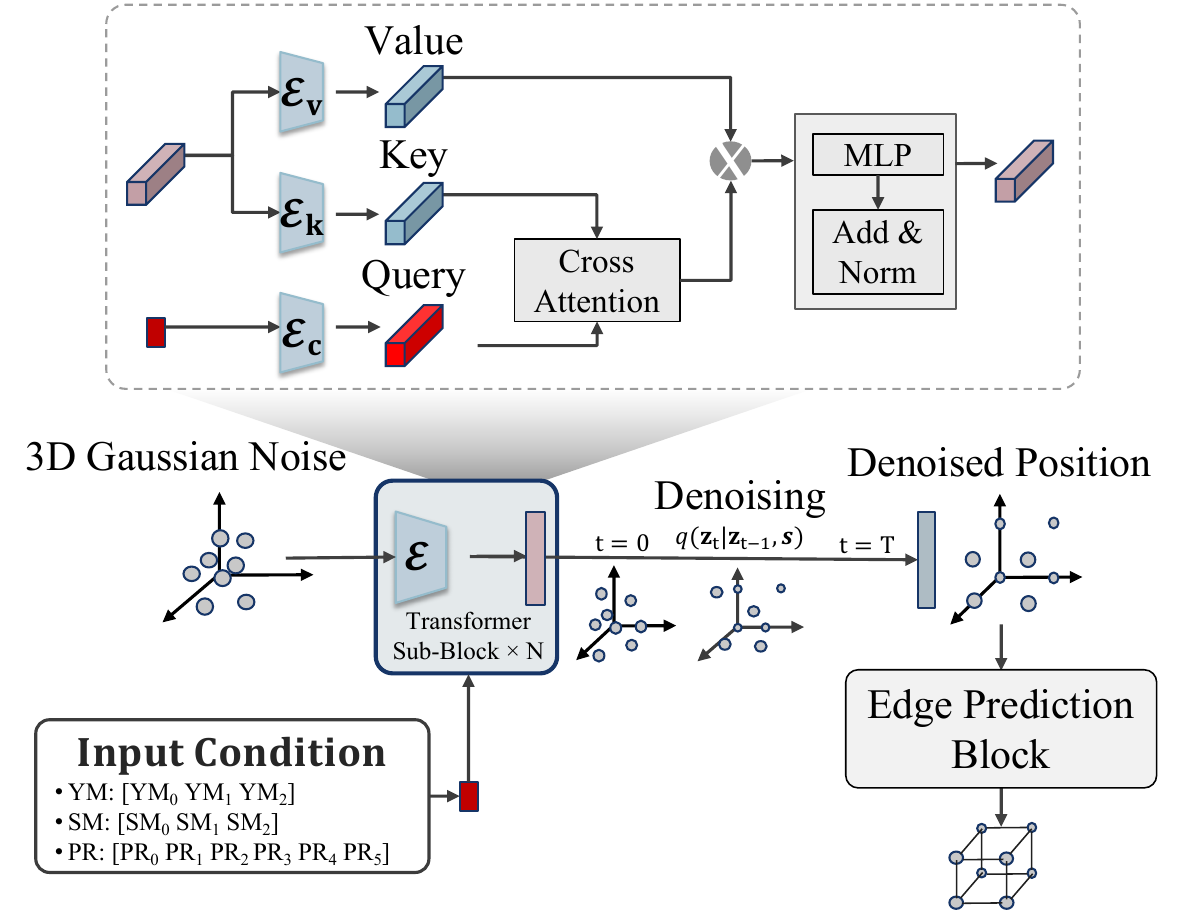}
    \vspace{-3mm}
	\caption{Illustration of 3D structure generation.}
    \vspace{-5mm}
\label{fig:3d_gen}
\end{figure}

\vspace{-2mm}
\paragraph{3D Structure Generation}
As shown in Figure~\ref{fig:3d_gen}, the initial 3D structure is generated with 
two components: (1) \textbf{Vertices Coordinate Diffusion}, which generates the 3D coordinates of the lattice vertices. It takes two inputs: random Gaussian noise and the lattice properties derived from the hypothesis generated in the preceding phase. The Gaussian noise, which matches the dimensionality of the 3D vertex coordinates, is input as a sequence of 3-dimensional tuples. The lattice properties are incorporated into the diffusion model via an attention mechanism implemented in the transformer sub-blocks. Specifically, the attention mechanism uses three components: query, key, and value. The query term is generated through a linear projection of the lattice properties, while the key and value terms are derived either from the input Gaussian noise or from the output of the previous transformer sub-block. Through iterative denoising, the diffusion block progressively refines the 3D coordinates, transforming the initial noise into structured vertex positions that align with the given lattice properties. (2)
\textbf{Lattice Edge Prediction}, which predicts edges between vertices. It 
uses a multi-layer perceptron (MLP) to evaluate the similarity between vertex representations and determine the presence or absence of edges. This module ensures the generated 3D lattice structure is cohesive and consistent with the hypothesized properties.



\vspace{-2mm}
\paragraph{3D Structure Refinement}
While the 3D lattice generation module can produce new and reasonable structures, it's still challenging to ensure the symmetric structure or periodic boundary conditions, which are crucial for improving mechanical properties of the metamaterial. 
Therefore, we introduce
a LLM-based structure refinement module. Inspired by \citet{gruver2024fine}, we feed the structure information as text into an LLM and refine
the lattice structure in two sequential stages: 
(1) \textbf{Node Refinement}, where the model optimizes nodes' symmetry and periodicity by adjusting node positions, adding necessary nodes, and removing redundant ones; and (2) \textbf{Edge Refinement},
where the model adjusts edge connections on the new node layout so that the overall structures satisfy the periodic boundary conditions necessary for fabrication.
Such a process is iterated for multiple cycles until the desired structure symmetry and periodicity are achieved simultaneously. 

\vspace{-1mm}
\section{Experiments and Evaluation}
\label{sec:case_study}
\vspace{-1mm}

\begin{figure*}[ht!]
    \centering
    \includegraphics[width=0.98\linewidth, trim={0 0 0 0},clip ]{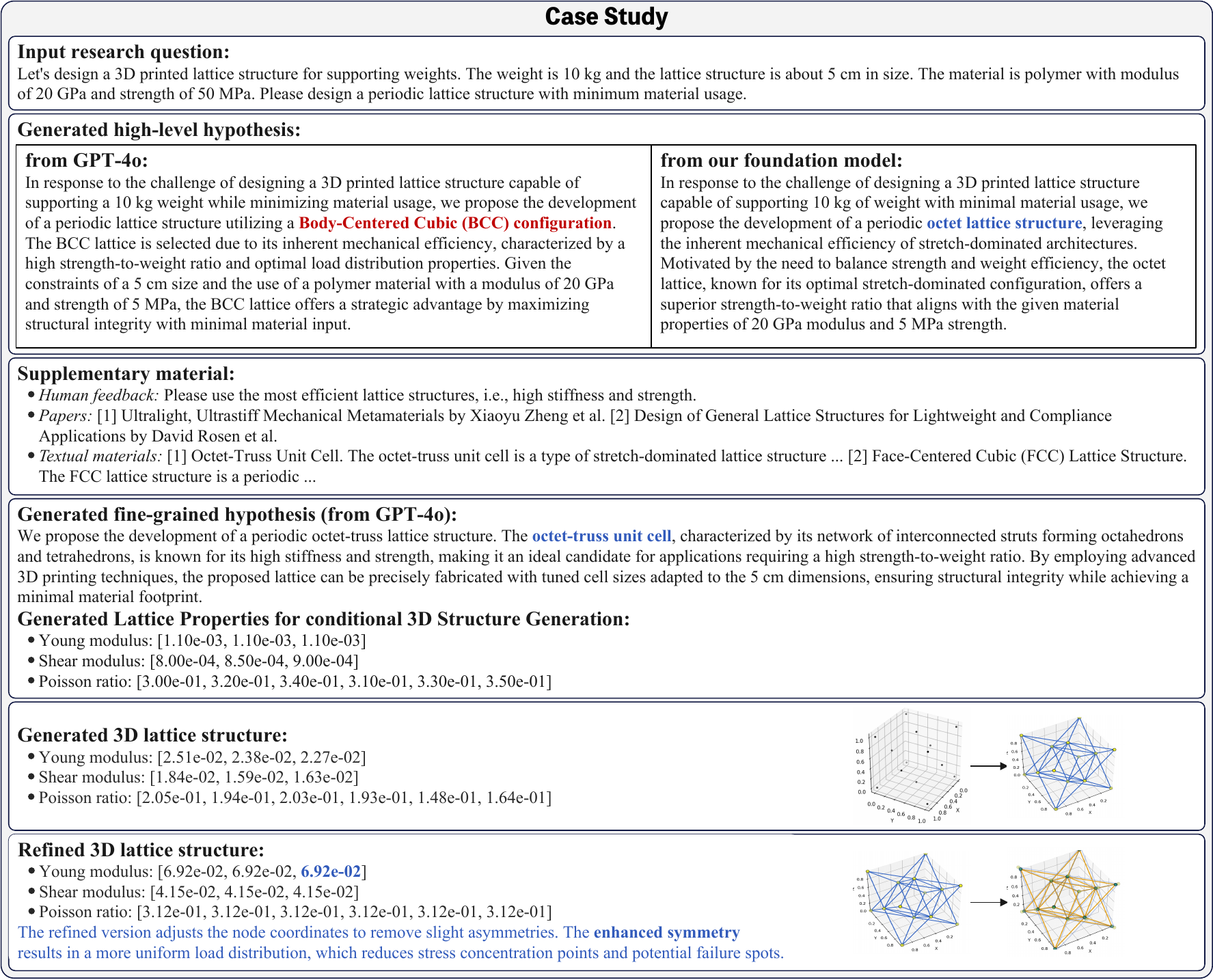}
    \vspace{-2mm}
	\caption{\textbf{Case study} of using \system{} to solve the weight minimization problem.}
    \vspace{-3mm}
\label{casestudy}
\end{figure*}

\paragraph{Case Study} 
To demonstrate the effectiveness of \system{},
we present a case study focusing on a classical weight minimization problem in materials science, as shown in Figure~\ref{casestudy}. 
The weight minimization problem represents an ideal test case for \system{} as it combines fundamental materials science challenges with clear industrial applications - designing structures that maximize support while minimizing material usage requires sophisticated reasoning about mechanical constraints while allowing for quantitative evaluation through well-established metrics. 
We present more additional case studies in Appendix~\ref{app:case_study}.

From the case study shown in Figure~\ref{casestudy}, we have the following observations. \textbf{First,} our system demonstrates superior performance in generating reasonable high-level hypotheses. For instance, when presented with the weight minimization problem, our fine-tuned foundation model directly proposed the octet lattice structure, recognizing its optimal stretch-dominated configuration and superior strength-to-weight ratio. In contrast, GPT-4o initially suggested a Body-Centered Cubic (BCC) configuration, which is less optimal for this specific application. This comparison highlights our model's enhanced domain knowledge and reasoning capabilities. 
\textbf{Second,} our system can effectively leverage human feedback and supplementary materials to improve hypothesis quality. When provided with expert feedback emphasizing ``\textit{use the most efficient lattice structures with high stiffness and strength}'' and relevant research papers (e.g., \citet{zheng2014ultralight} and \citet{deshpande2001effective}), GPT-4o successfully revised its hypothesis from BCC to the more suitable octet-truss structure. The retrieved papers and text materials can also provide important references for our system to predict reasonable and valid ranges for the lattice properties in the fine-grained hypothesis.
\textbf{Third,} we demonstrate that our system can generate valid 3D lattice structure and our refinement model can effectively refine and improve the generated structure. From Figure~\ref{casestudy}, the initial generated structure, while reasonable, showed slight asymmetries in its lattice properties. Through our refinement process, the structure was optimized to achieve better cubic symmetry with improved Young's modulus in the $z$ axis. These improvements enhanced load distribution and reduced stress concentration points, resulting in superior overall strength and stiffness in the final design.

\begin{table}[t]
\centering
\resizebox{0.37\textwidth}{!}{
\begin{tabular}{lcccc}
\toprule
              & \multicolumn{2}{c}{\textbf{Novelty}}        & \multicolumn{2}{c}{\textbf{Feasibility}}    \\
              & \textbf{Mean}      & \textbf{Std}       & \textbf{Mean}      & \textbf{Std}       \\ \midrule
GPT-4o        & 0.803                                                            & 0.481                                                           & \textbf{1.591}                                                                & 0.605                                                               \\
\midrule
\system{} & \multicolumn{1}{l}{}                                              & \multicolumn{1}{l}{}                                             & \multicolumn{1}{l}{}                                                  & \multicolumn{1}{l}{}                                                 \\
\hspace{1mm} - w/o Human     & 0.833                                                            & 0.582                                                           & 1.227                                                               & 0.651                                                             \\
\hspace{1mm} - w/ Human      & \textbf{1.712}                                                           & 0.424                                                           & 1.47                                                            & 0.499          \\ \bottomrule                                                    
\end{tabular}
}
\vspace{-1mm}
\caption{Hypothesis Evaluation for \textbf{Novelty} and \textbf{Feasibility}. 
Each evaluation aspect has three scales, where score 0 is the worst and score 2 is the best.}
\vspace{-2mm}
\label{tab:human_eval}
\end{table}

\begin{table}[t]
\centering
\resizebox{0.47\textwidth}{!}{
\begin{tabular}{c|ccc}
\toprule
\textbf{Category} & \textbf{Error Threshold} & \textbf{Intra-cell Validity} & \textbf{Inter-cell Validity} \\ 
\midrule
\multirow{4}{*}{\begin{tabular}[c]{@{}c@{}}Initial\\ Lattice\end{tabular}} 
& 0.005 & 22  & 0   \\ 
& 0.01  & 93  & 41  \\ 
& 0.02  & 97  & 82  \\ 
& 0.04  & 98  & 98  \\ 
\midrule
\multirow{4}{*}{\begin{tabular}[c]{@{}c@{}}Refined\\ Lattice\end{tabular}} 
& 0.005 & 55   & 39   \\ 
& 0.01  & 94   & 52   \\  
& 0.02  & 98  & 83  \\ 
& 0.04  & 98 & 98  \\ 
\bottomrule
\end{tabular}}
\vspace{-2mm}
\caption{\textbf{Validity (\%)} of the initially generated lattice structure and refined lattice structure.}
\label{tab:3d_gen}
\vspace{-5mm}
\end{table}

\vspace{-2mm}
\paragraph{Quantitative Evaluation} 
We also conduct quantitative experiments to evaluate textual hypothesis generation in Table~\ref{tab:human_eval}, and 3D structure generation and refinement in Table~\ref{tab:3d_gen}.
For hypothesis generation evaluation, we employ 
two widely adopted aspects, i.e., \textit{novelty} and \textit{feasibility}~\cite{si2024can}. 
We test our system on 11 metamaterial research questions carefully curated by domain experts, and ask three senior researchers (i.e., 1 professor and 2 postdocs) in material science to evaluate the hypotheses generated by GPT-4o and our \system{} with or without human feedback. 
We include more details including evaluation criteria and the results for each instance in Appendix~\ref{app:hypogen_eval}. From the results in Table~\ref{tab:human_eval} and the comments from the human evaluators, we have the following findings: 
(1) The novelty and feasibility of the hypotheses generated by \system{} can be significantly improved after incorporating human feedback. Our approach outperforms the GPT-4o baseline in terms of novelty by a substantial margin while maintaining a competitive result in feasibility.
(2) GPT-4o's hypotheses, while easier to complete and thus having a higher feasibility score, tend to be more trivial and lack the depth required for truly novel or ambitious ideas. In contrast, \system{} with human feedback produces hypotheses that are more complex and innovative, which inherently makes them more challenging to implement and leads to a slightly lower feasibility score.

For 3D structure validation, we design two metrics: \textit{intra-cell validity} and \textit{inter-cell validity}. There are two requirements for a structure to be intra-cell valid: all vertices should be connected and the structure should be symmetric, which means for each vertex, there should be a counterpart vertex for it, to make up a symmetric pair. To assert whether a structure is inter-cell valid, we first find a cubic frame corresponding to the unit cell of the structure, and then test whether any vertices exceed the boundary of the cubic frame. If not, the structure is considered to be inter-cell valid.
Since the structures generated or refined are not perfectly precise, we allow the vertices to deviate slightly from their supposed location. A series of error thresholds (from 0.005 to 0.04) are set for the evaluation of structures. For each threshold, if a vertex suffers a deviation of more than the cell size times the threshold, then the structure is determined to be intra-cell or inter-call invalid. From Table~\ref{tab:3d_gen} it can be observed that as the threshold increases, the validity also increases. If a relative error of 0.03 is allowed, then the majority of generated structures will be valid. Refined structures also show a similar pattern, but convergence will be reached with a lower threshold, which verifies the effectiveness of the refinement step.




\vspace{-1mm}
\section{Related Works}
\label{sec:related_work}
\vspace{-1mm}

Automated scientific discovery has progressed significantly
given the recent advancements of large foundation models~\cite{wang2023scientific, lu2024ai, papadimitriou2024ai, xiong2024improving, ishikawa2024hypothesis}, 
such as logical reasoning~\cite{sun2024scieval}, multi-agent collaboration~\cite{ma2024sciagent,jansen2024discoveryworld,baek2024researchagent,arawjo2024chainforge}, tool and retrieval-based augmentation~\cite{prince2024opportunities,qi2024rora,huang2024crispr}. 
However, the application of AI and LLMs for autonomous novel material discovery faces several challenges, including high error rates and hallucinations~\cite{lehr2024chatgpt,miret2024llms}, lacking  interpretability~\cite{liu2024beyond,lei2024materials}, rigorous logical and mathematical reasoning 
capabilities~\cite{park2024can,deb2024chatgpt,zhang2024honeycomb}, leading to scientifically unsound or unverifiable hypotheses. 
Additionally, LLMs often struggle to extrapolate beyond existing knowledge due to the nature of their 
pre-training objective~\cite{petroni2019language, achiam2023gpt}. 
To address these challenges, \system{} employs a human-in-the-loop approach to integrate expert feedback and domain-specific knowledge into the design of novel materials, which significantly mitigates hallucinations and aligns outputs with scientific principles.
\vspace{-1mm}
\section{Conclusion}
\label{sec:conclusion}
\vspace{-1mm}

We propose \system{} 
that accelerates the discovery of novel metamaterials by integrating advanced AI capabilities with expert oversight. Experiments and human evaluation demonstrate that \system{} can generate reasonable and valid metamaterial design given the user's research question. 
Future works include improving the validity of lattice properties using more advanced retrieval techniques and building more powerful models to generate 3D metamaterial structures in more diverse scenarios. We believe our work has significant potential to advance the field of metamaterial design and 
other broad scientific domains.

\section*{Acknowledgments}
We thank Dr. Hongshun Chen from the University of Pennsylvania for the help and support on hypothesis evaluation. This research is supported by the U.S. DARPA FoundSci Program \#HR00112490370. The views and conclusions contained herein are those of the authors and should not be interpreted as necessarily representing the official policies, either expressed or implied, of the U.S. Government. The U.S. Government is authorized to reproduce and distribute reprints for governmental purposes notwithstanding any copyright annotation therein.

\bibliography{custom}

\begin{thebibliography}{43}
\providecommand{\natexlab}[1]{#1}

\bibitem[{Achiam et~al.(2023)Achiam, Adler, Agarwal, Ahmad, Akkaya, Aleman, Almeida, Altenschmidt, Altman, Anadkat et~al.}]{achiam2023gpt}
Josh Achiam, Steven Adler, Sandhini Agarwal, Lama Ahmad, Ilge Akkaya, Florencia~Leoni Aleman, Diogo Almeida, Janko Altenschmidt, Sam Altman, Shyamal Anadkat, et~al. 2023.
\newblock Gpt-4 technical report.
\newblock \emph{arXiv preprint arXiv:2303.08774}.

\bibitem[{Arawjo et~al.(2024)Arawjo, Swoopes, Vaithilingam, Wattenberg, and Glassman}]{arawjo2024chainforge}
Ian Arawjo, Chelse Swoopes, Priyan Vaithilingam, Martin Wattenberg, and Elena~L Glassman. 2024.
\newblock Chainforge: A visual toolkit for prompt engineering and llm hypothesis testing.
\newblock In \emph{Proceedings of the CHI Conference on Human Factors in Computing Systems}, pages 1--18.

\bibitem[{Baek et~al.(2024)Baek, Jauhar, Cucerzan, and Hwang}]{baek2024researchagent}
Jinheon Baek, Sujay~Kumar Jauhar, Silviu Cucerzan, and Sung~Ju Hwang. 2024.
\newblock Researchagent: Iterative research idea generation over scientific literature with large language models.
\newblock \emph{arXiv preprint arXiv:2404.07738}.

\bibitem[{Bauer et~al.(2017)Bauer, Meza, Schaedler, Schwaiger, Zheng, and Valdevit}]{bauer2017nanolattices}
Jens Bauer, Lucas~R Meza, Tobias~A Schaedler, Ruth Schwaiger, Xiaoyu Zheng, and Lorenzo Valdevit. 2017.
\newblock Nanolattices: an emerging class of mechanical metamaterials.
\newblock \emph{Advanced Materials}, 29(40):1701850.

\bibitem[{Bertoldi et~al.(2017)Bertoldi, Vitelli, Christensen, and Van~Hecke}]{bertoldi2017flexible}
Katia Bertoldi, Vincenzo Vitelli, Johan Christensen, and Martin Van~Hecke. 2017.
\newblock Flexible mechanical metamaterials.
\newblock \emph{Nature Reviews Materials}, 2(11):1--11.

\bibitem[{Deb et~al.(2024)Deb, Saikia, Dihingia, and Sastry}]{deb2024chatgpt}
Jyotirmoy Deb, Lakshi Saikia, Kripa~Dristi Dihingia, and G~Narahari Sastry. 2024.
\newblock Chatgpt in the material design: Selected case studies to assess the potential of chatgpt.
\newblock \emph{Journal of Chemical Information and Modeling}, 64(3):799--811.

\bibitem[{Deshpande et~al.(2001)Deshpande, Fleck, and Ashby}]{deshpande2001effective}
Vikram~S Deshpande, Norman~A Fleck, and Michael~F Ashby. 2001.
\newblock Effective properties of the octet-truss lattice material.
\newblock \emph{Journal of the Mechanics and Physics of Solids}, 49(8):1747--1769.

\bibitem[{Dubey et~al.(2024)Dubey, Jauhri, Pandey, Kadian, Al-Dahle, Letman, Mathur, Schelten, Yang, Fan et~al.}]{dubey2024llama}
Abhimanyu Dubey, Abhinav Jauhri, Abhinav Pandey, Abhishek Kadian, Ahmad Al-Dahle, Aiesha Letman, Akhil Mathur, Alan Schelten, Amy Yang, Angela Fan, et~al. 2024.
\newblock The llama 3 herd of models.
\newblock \emph{arXiv preprint arXiv:2407.21783}.

\bibitem[{Gao et~al.(2021)Gao, Yao, and Chen}]{gao-etal-2021-simcse}
Tianyu Gao, Xingcheng Yao, and Danqi Chen. 2021.
\newblock \href {https://doi.org/10.18653/v1/2021.emnlp-main.552} {{S}im{CSE}: Simple contrastive learning of sentence embeddings}.
\newblock In \emph{Proceedings of the 2021 Conference on Empirical Methods in Natural Language Processing}, pages 6894--6910, Online and Punta Cana, Dominican Republic. Association for Computational Linguistics.

\bibitem[{Gruver et~al.(2024)Gruver, Sriram, Madotto, Wilson, Zitnick, and Ulissi}]{gruver2024fine}
Nate Gruver, Anuroop Sriram, Andrea Madotto, Andrew~Gordon Wilson, C~Lawrence Zitnick, and Zachary Ulissi. 2024.
\newblock Fine-tuned language models generate stable inorganic materials as text.
\newblock \emph{arXiv preprint arXiv:2402.04379}.

\bibitem[{Hong(2023)}]{hong2023chatgpt}
Zijian Hong. 2023.
\newblock Chatgpt for computational materials science: A perspective.
\newblock \emph{Energy Material Advances}, 4:0026.

\bibitem[{Hu et~al.(2022)Hu, yelong shen, Wallis, Allen-Zhu, Li, Wang, Wang, and Chen}]{hu2022lora}
Edward~J Hu, yelong shen, Phillip Wallis, Zeyuan Allen-Zhu, Yuanzhi Li, Shean Wang, Lu~Wang, and Weizhu Chen. 2022.
\newblock \href {https://openreview.net/forum?id=nZeVKeeFYf9} {Lo{RA}: Low-rank adaptation of large language models}.
\newblock In \emph{International Conference on Learning Representations}.

\bibitem[{Huang et~al.(2024)Huang, Qu, Cousins, Johnson, Yin, Shah, Zhou, Altman, Wang, and Cong}]{huang2024crispr}
Kaixuan Huang, Yuanhao Qu, Henry Cousins, William~A Johnson, Di~Yin, Mihir Shah, Denny Zhou, Russ Altman, Mengdi Wang, and Le~Cong. 2024.
\newblock Crispr-gpt: An llm agent for automated design of gene-editing experiments.
\newblock \emph{arXiv preprint arXiv:2404.18021}.

\bibitem[{Ishikawa(2024)}]{ishikawa2024hypothesis}
Hiroshi Ishikawa. 2024.
\newblock \emph{Hypothesis Generation and Interpretation: Design Principles and Patterns for Big Data Applications}, volume 139.
\newblock Springer Nature.

\bibitem[{Jansen et~al.(2024)Jansen, C{\^o}t{\'e}, Khot, Bransom, Mishra, Majumder, Tafjord, and Clark}]{jansen2024discoveryworld}
Peter Jansen, Marc-Alexandre C{\^o}t{\'e}, Tushar Khot, Erin Bransom, Bhavana~Dalvi Mishra, Bodhisattwa~Prasad Majumder, Oyvind Tafjord, and Peter Clark. 2024.
\newblock Discoveryworld: A virtual environment for developing and evaluating automated scientific discovery agents.
\newblock \emph{arXiv preprint arXiv:2406.06769}.

\bibitem[{Jia et~al.(2020)Jia, Liu, Jiang, and Wang}]{jia2020engineering}
Zian Jia, Fan Liu, Xihang Jiang, and Lifeng Wang. 2020.
\newblock Engineering lattice metamaterials for extreme property, programmability, and multifunctionality.
\newblock \emph{Journal of Applied Physics}, 127(15).

\bibitem[{Jiao and Alavi(2021)}]{jiao2021artificial}
Pengcheng Jiao and Amir~H Alavi. 2021.
\newblock Artificial intelligence-enabled smart mechanical metamaterials: advent and future trends.
\newblock \emph{International Materials Reviews}, 66(6):365--393.

\bibitem[{Kadic et~al.(2019)Kadic, Milton, van Hecke, and Wegener}]{kadic20193d}
Muamer Kadic, Graeme~W Milton, Martin van Hecke, and Martin Wegener. 2019.
\newblock 3d metamaterials.
\newblock \emph{Nature Reviews Physics}, 1(3):198--210.

\bibitem[{Khattab and Zaharia(2020)}]{ColBERT}
Omar Khattab and Matei Zaharia. 2020.
\newblock \href {https://doi.org/10.1145/3397271.3401075} {Colbert: Efficient and effective passage search via contextualized late interaction over bert}.
\newblock In \emph{Proceedings of the 43rd International ACM SIGIR Conference on Research and Development in Information Retrieval}, SIGIR '20, page 39–48, New York, NY, USA. Association for Computing Machinery.

\bibitem[{Kingma(2014)}]{kingma2014adam}
Diederik~P Kingma. 2014.
\newblock Adam: A method for stochastic optimization.
\newblock \emph{arXiv preprint arXiv:1412.6980}.

\bibitem[{Lehr et~al.(2024)Lehr, Caliskan, Liyanage, and Banaji}]{lehr2024chatgpt}
Steven~A Lehr, Aylin Caliskan, Suneragiri Liyanage, and Mahzarin~R Banaji. 2024.
\newblock Chatgpt as research scientist: Probing gpt’s capabilities as a research librarian, research ethicist, data generator, and data predictor.
\newblock \emph{Proceedings of the National Academy of Sciences}, 121(35):e2404328121.

\bibitem[{Lei et~al.(2024)Lei, Docherty, and Cooper}]{lei2024materials}
Ge~Lei, Ronan Docherty, and Samuel~J Cooper. 2024.
\newblock Materials science in the era of large language models: a perspective.
\newblock \emph{Digital Discovery}.

\bibitem[{Liu et~al.(2024)Liu, Polak, Kim, Shuvo, Deodhar, Han, Morgan, and Oh}]{liu2024beyond}
Quanliang Liu, Maciej~P Polak, So~Yeon Kim, MD~Shuvo, Hrishikesh~Shridhar Deodhar, Jeongsoo Han, Dane Morgan, and Hyunseok Oh. 2024.
\newblock Beyond designer's knowledge: Generating materials design hypotheses via large language models.
\newblock \emph{arXiv preprint arXiv:2409.06756}.

\bibitem[{Lu et~al.(2024)Lu, Lu, Lange, Foerster, Clune, and Ha}]{lu2024ai}
Chris Lu, Cong Lu, Robert~Tjarko Lange, Jakob Foerster, Jeff Clune, and David Ha. 2024.
\newblock The ai scientist: Towards fully automated open-ended scientific discovery.
\newblock \emph{arXiv preprint arXiv:2408.06292}.

\bibitem[{Lumpe and Stankovic(2021)}]{lumpe2021exploring}
Thomas~S Lumpe and Tino Stankovic. 2021.
\newblock Exploring the property space of periodic cellular structures based on crystal networks.
\newblock \emph{Proceedings of the National Academy of Sciences}, 118(7):e2003504118.

\bibitem[{Ma et~al.(2024)Ma, Gou, Hao, Xu, Wang, Pan, Yang, Cao, Sun, Awadalla et~al.}]{ma2024sciagent}
Yubo Ma, Zhibin Gou, Junheng Hao, Ruochen Xu, Shuohang Wang, Liangming Pan, Yujiu Yang, Yixin Cao, Aixin Sun, Hany Awadalla, et~al. 2024.
\newblock Sciagent: Tool-augmented language models for scientific reasoning.
\newblock \emph{arXiv preprint arXiv:2402.11451}.

\bibitem[{Miret and Krishnan(2024)}]{miret2024llms}
Santiago Miret and NM~Krishnan. 2024.
\newblock Are llms ready for real-world materials discovery?
\newblock \emph{arXiv preprint arXiv:2402.05200}.

\bibitem[{OpenAI et~al.(2024)OpenAI, :, Hurst, Lerer, Goucher, Perelman, Ramesh, Clark, Ostrow, Welihinda, Hayes, Radford, Mądry, Baker-Whitcomb, Beutel, Borzunov, Carney, Chow, Kirillov, Nichol, Paino, Renzin, Passos, Kirillov, Christakis, Conneau, Kamali, Jabri, Moyer, Tam, Crookes, Tootoochian, Tootoonchian, Kumar, Vallone, Karpathy, Braunstein, Cann, Codispoti, Galu, Kondrich, Tulloch, Mishchenko, Baek, Jiang, Pelisse, Woodford, Gosalia, Dhar, Pantuliano, Nayak, Oliver, Zoph, Ghorbani, Leimberger, Rossen, Sokolowsky, Wang, Zweig, Hoover, Samic, McGrew, Spero, Giertler, Cheng, Lightcap, Walkin, Quinn, Guarraci, Hsu, Kellogg, Eastman, Lugaresi, Wainwright, Bassin, Hudson, Chu, Nelson, Li, Shern, Conger, Barette, Voss, Ding, Lu, Zhang, Beaumont, Hallacy, Koch, Gibson, Kim, Choi, McLeavey, Hesse, Fischer, Winter, Czarnecki, Jarvis, Wei, Koumouzelis, Sherburn, Kappler, Levin, Levy, Carr, Farhi, Mely, Robinson, Sasaki, Jin, Valladares, Tsipras, Li, Nguyen, Findlay, Oiwoh, Wong, Asdar, Proehl, Yang, Antonow,
  Kramer, Peterson, Sigler, Wallace, Brevdo, Mays, Khorasani, Such, Raso, Zhang, von Lohmann, Sulit, Goh, Oden, Salmon, Starace, Brockman, Salman, Bao, Hu, Wong, Wang, Schmidt, Whitney, Jun, Kirchner, de~Oliveira~Pinto, Ren, Chang, Chung, Kivlichan, O'Connell, O'Connell, Osband, Silber, Sohl, Okuyucu, Lan, Kostrikov, Sutskever, Kanitscheider, Gulrajani, Coxon, Menick, Pachocki, Aung, Betker, Crooks, Lennon, Kiros, Leike, Park, Kwon, Phang, Teplitz, Wei, Wolfe, Chen, Harris, Varavva, Lee, Shieh, Lin, Yu, Weng, Tang, Yu, Jang, Candela, Beutler, Landers, Parish, Heidecke, Schulman, Lachman, McKay, Uesato, Ward, Kim, Huizinga, Sitkin, Kraaijeveld, Gross, Kaplan, Snyder, Achiam, Jiao, Lee, Zhuang, Harriman, Fricke, Hayashi, Singhal, Shi, Karthik, Wood, Rimbach, Hsu, Nguyen, Gu-Lemberg, Button, Liu, Howe, Muthukumar, Luther, Ahmad, Kai, Itow, Workman, Pathak, Chen, Jing, Guy, Fedus, Zhou, Mamitsuka, Weng, McCallum, Held, Ouyang, Feuvrier, Zhang, Kondraciuk, Kaiser, Hewitt, Metz, Doshi, Aflak, Simens, Boyd,
  Thompson, Dukhan, Chen, Gray, Hudnall, Zhang, Aljubeh, Litwin, Zeng, Johnson, Shetty, Gupta, Shah, Yatbaz, Yang, Zhong, Glaese, Chen, Janner, Lampe, Petrov, Wu, Wang, Fradin, Pokrass, Castro, de~Castro, Pavlov, Brundage, Wang, Khan, Murati, Bavarian, Lin, Yesildal, Soto, Gimelshein, Cone, Staudacher, Summers, LaFontaine, Chowdhury, Ryder, Stathas, Turley, Tezak, Felix, Kudige, Keskar, Deutsch, Bundick, Puckett, Nachum, Okelola, Boiko, Murk, Jaffe, Watkins, Godement, Campbell-Moore, Chao, McMillan, Belov, Su, Bak, Bakkum, Deng, Dolan, Hoeschele, Welinder, Tillet, Pronin, Tillet, Dhariwal, Yuan, Dias, Lim, Arora, Troll, Lin, Lopes, Puri, Miyara, Leike, Gaubert, Zamani, Wang, Donnelly, Honsby, Smith, Sahai, Ramchandani, Huet, Carmichael, Zellers, Chen, Chen, Nigmatullin, Cheu, Jain, Altman, Schoenholz, Toizer, Miserendino, Agarwal, Culver, Ethersmith, Gray, Grove, Metzger, Hermani, Jain, Zhao, Wu, Jomoto, Wu, Shuaiqi, Xia, Phene, Papay, Narayanan, Coffey, Lee, Hall, Balaji, Broda, Stramer, Xu, Gogineni,
  Christianson, Sanders, Patwardhan, Cunninghman, Degry, Dimson, Raoux, Shadwell, Zheng, Underwood, Markov, Sherbakov, Rubin, Stasi, Kaftan, Heywood, Peterson, Walters, Eloundou, Qi, Moeller, Monaco, Kuo, Fomenko, Chang, Zheng, Zhou, Manassra, Sheu, Zaremba, Patil, Qian, Kim, Cheng, Zhang, He, Zhang, Jin, Dai, and Malkov}]{openai2024gpt4ocard}
OpenAI, :, Aaron Hurst, Adam Lerer, Adam~P. Goucher, Adam Perelman, Aditya Ramesh, Aidan Clark, AJ~Ostrow, Akila Welihinda, Alan Hayes, Alec Radford, Aleksander Mądry, Alex Baker-Whitcomb, Alex Beutel, Alex Borzunov, Alex Carney, Alex Chow, Alex Kirillov, Alex Nichol, Alex Paino, Alex Renzin, Alex~Tachard Passos, Alexander Kirillov, Alexi Christakis, Alexis Conneau, Ali Kamali, Allan Jabri, Allison Moyer, Allison Tam, Amadou Crookes, Amin Tootoochian, Amin Tootoonchian, Ananya Kumar, Andrea Vallone, Andrej Karpathy, Andrew Braunstein, Andrew Cann, Andrew Codispoti, Andrew Galu, Andrew Kondrich, Andrew Tulloch, Andrey Mishchenko, Angela Baek, Angela Jiang, Antoine Pelisse, Antonia Woodford, Anuj Gosalia, Arka Dhar, Ashley Pantuliano, Avi Nayak, Avital Oliver, Barret Zoph, Behrooz Ghorbani, Ben Leimberger, Ben Rossen, Ben Sokolowsky, Ben Wang, Benjamin Zweig, Beth Hoover, Blake Samic, Bob McGrew, Bobby Spero, Bogo Giertler, Bowen Cheng, Brad Lightcap, Brandon Walkin, Brendan Quinn, Brian Guarraci, Brian Hsu,
  Bright Kellogg, Brydon Eastman, Camillo Lugaresi, Carroll Wainwright, Cary Bassin, Cary Hudson, Casey Chu, Chad Nelson, Chak Li, Chan~Jun Shern, Channing Conger, Charlotte Barette, Chelsea Voss, Chen Ding, Cheng Lu, Chong Zhang, Chris Beaumont, Chris Hallacy, Chris Koch, Christian Gibson, Christina Kim, Christine Choi, Christine McLeavey, Christopher Hesse, Claudia Fischer, Clemens Winter, Coley Czarnecki, Colin Jarvis, Colin Wei, Constantin Koumouzelis, Dane Sherburn, Daniel Kappler, Daniel Levin, Daniel Levy, David Carr, David Farhi, David Mely, David Robinson, David Sasaki, Denny Jin, Dev Valladares, Dimitris Tsipras, Doug Li, Duc~Phong Nguyen, Duncan Findlay, Edede Oiwoh, Edmund Wong, Ehsan Asdar, Elizabeth Proehl, Elizabeth Yang, Eric Antonow, Eric Kramer, Eric Peterson, Eric Sigler, Eric Wallace, Eugene Brevdo, Evan Mays, Farzad Khorasani, Felipe~Petroski Such, Filippo Raso, Francis Zhang, Fred von Lohmann, Freddie Sulit, Gabriel Goh, Gene Oden, Geoff Salmon, Giulio Starace, Greg Brockman, Hadi
  Salman, Haiming Bao, Haitang Hu, Hannah Wong, Haoyu Wang, Heather Schmidt, Heather Whitney, Heewoo Jun, Hendrik Kirchner, Henrique~Ponde de~Oliveira~Pinto, Hongyu Ren, Huiwen Chang, Hyung~Won Chung, Ian Kivlichan, Ian O'Connell, Ian O'Connell, Ian Osband, Ian Silber, Ian Sohl, Ibrahim Okuyucu, Ikai Lan, Ilya Kostrikov, Ilya Sutskever, Ingmar Kanitscheider, Ishaan Gulrajani, Jacob Coxon, Jacob Menick, Jakub Pachocki, James Aung, James Betker, James Crooks, James Lennon, Jamie Kiros, Jan Leike, Jane Park, Jason Kwon, Jason Phang, Jason Teplitz, Jason Wei, Jason Wolfe, Jay Chen, Jeff Harris, Jenia Varavva, Jessica~Gan Lee, Jessica Shieh, Ji~Lin, Jiahui Yu, Jiayi Weng, Jie Tang, Jieqi Yu, Joanne Jang, Joaquin~Quinonero Candela, Joe Beutler, Joe Landers, Joel Parish, Johannes Heidecke, John Schulman, Jonathan Lachman, Jonathan McKay, Jonathan Uesato, Jonathan Ward, Jong~Wook Kim, Joost Huizinga, Jordan Sitkin, Jos Kraaijeveld, Josh Gross, Josh Kaplan, Josh Snyder, Joshua Achiam, Joy Jiao, Joyce Lee, Juntang
  Zhuang, Justyn Harriman, Kai Fricke, Kai Hayashi, Karan Singhal, Katy Shi, Kavin Karthik, Kayla Wood, Kendra Rimbach, Kenny Hsu, Kenny Nguyen, Keren Gu-Lemberg, Kevin Button, Kevin Liu, Kiel Howe, Krithika Muthukumar, Kyle Luther, Lama Ahmad, Larry Kai, Lauren Itow, Lauren Workman, Leher Pathak, Leo Chen, Li~Jing, Lia Guy, Liam Fedus, Liang Zhou, Lien Mamitsuka, Lilian Weng, Lindsay McCallum, Lindsey Held, Long Ouyang, Louis Feuvrier, Lu~Zhang, Lukas Kondraciuk, Lukasz Kaiser, Luke Hewitt, Luke Metz, Lyric Doshi, Mada Aflak, Maddie Simens, Madelaine Boyd, Madeleine Thompson, Marat Dukhan, Mark Chen, Mark Gray, Mark Hudnall, Marvin Zhang, Marwan Aljubeh, Mateusz Litwin, Matthew Zeng, Max Johnson, Maya Shetty, Mayank Gupta, Meghan Shah, Mehmet Yatbaz, Meng~Jia Yang, Mengchao Zhong, Mia Glaese, Mianna Chen, Michael Janner, Michael Lampe, Michael Petrov, Michael Wu, Michele Wang, Michelle Fradin, Michelle Pokrass, Miguel Castro, Miguel Oom~Temudo de~Castro, Mikhail Pavlov, Miles Brundage, Miles Wang, Minal
  Khan, Mira Murati, Mo~Bavarian, Molly Lin, Murat Yesildal, Nacho Soto, Natalia Gimelshein, Natalie Cone, Natalie Staudacher, Natalie Summers, Natan LaFontaine, Neil Chowdhury, Nick Ryder, Nick Stathas, Nick Turley, Nik Tezak, Niko Felix, Nithanth Kudige, Nitish Keskar, Noah Deutsch, Noel Bundick, Nora Puckett, Ofir Nachum, Ola Okelola, Oleg Boiko, Oleg Murk, Oliver Jaffe, Olivia Watkins, Olivier Godement, Owen Campbell-Moore, Patrick Chao, Paul McMillan, Pavel Belov, Peng Su, Peter Bak, Peter Bakkum, Peter Deng, Peter Dolan, Peter Hoeschele, Peter Welinder, Phil Tillet, Philip Pronin, Philippe Tillet, Prafulla Dhariwal, Qiming Yuan, Rachel Dias, Rachel Lim, Rahul Arora, Rajan Troll, Randall Lin, Rapha~Gontijo Lopes, Raul Puri, Reah Miyara, Reimar Leike, Renaud Gaubert, Reza Zamani, Ricky Wang, Rob Donnelly, Rob Honsby, Rocky Smith, Rohan Sahai, Rohit Ramchandani, Romain Huet, Rory Carmichael, Rowan Zellers, Roy Chen, Ruby Chen, Ruslan Nigmatullin, Ryan Cheu, Saachi Jain, Sam Altman, Sam Schoenholz, Sam
  Toizer, Samuel Miserendino, Sandhini Agarwal, Sara Culver, Scott Ethersmith, Scott Gray, Sean Grove, Sean Metzger, Shamez Hermani, Shantanu Jain, Shengjia Zhao, Sherwin Wu, Shino Jomoto, Shirong Wu, Shuaiqi, Xia, Sonia Phene, Spencer Papay, Srinivas Narayanan, Steve Coffey, Steve Lee, Stewart Hall, Suchir Balaji, Tal Broda, Tal Stramer, Tao Xu, Tarun Gogineni, Taya Christianson, Ted Sanders, Tejal Patwardhan, Thomas Cunninghman, Thomas Degry, Thomas Dimson, Thomas Raoux, Thomas Shadwell, Tianhao Zheng, Todd Underwood, Todor Markov, Toki Sherbakov, Tom Rubin, Tom Stasi, Tomer Kaftan, Tristan Heywood, Troy Peterson, Tyce Walters, Tyna Eloundou, Valerie Qi, Veit Moeller, Vinnie Monaco, Vishal Kuo, Vlad Fomenko, Wayne Chang, Weiyi Zheng, Wenda Zhou, Wesam Manassra, Will Sheu, Wojciech Zaremba, Yash Patil, Yilei Qian, Yongjik Kim, Youlong Cheng, Yu~Zhang, Yuchen He, Yuchen Zhang, Yujia Jin, Yunxing Dai, and Yury Malkov. 2024.
\newblock \href {https://arxiv.org/abs/2410.21276} {Gpt-4o system card}.
\newblock \emph{Preprint}, arXiv:2410.21276.

\bibitem[{Papadimitriou et~al.(2024)Papadimitriou, Gialampoukidis, Vrochidis, and Kompatsiaris}]{papadimitriou2024ai}
Ioannis Papadimitriou, Ilias Gialampoukidis, Stefanos Vrochidis, and Ioannis Kompatsiaris. 2024.
\newblock Ai methods in materials design, discovery and manufacturing: A review.
\newblock \emph{Computational Materials Science}, 235:112793.

\bibitem[{Park et~al.(2024)Park, Kaplan, Ren, Hsu, Li, Xu, Li, and Li}]{park2024can}
Yang~Jeong Park, Daniel Kaplan, Zhichu Ren, Chia-Wei Hsu, Changhao Li, Haowei Xu, Sipei Li, and Ju~Li. 2024.
\newblock Can chatgpt be used to generate scientific hypotheses?
\newblock \emph{Journal of Materiomics}, 10(3):578--584.

\bibitem[{Petroni et~al.(2019)Petroni, Rockt{\"a}schel, Lewis, Bakhtin, Wu, Miller, and Riedel}]{petroni2019language}
Fabio Petroni, Tim Rockt{\"a}schel, Patrick Lewis, Anton Bakhtin, Yuxiang Wu, Alexander~H Miller, and Sebastian Riedel. 2019.
\newblock Language models as knowledge bases?
\newblock \emph{arXiv preprint arXiv:1909.01066}.

\bibitem[{Prince et~al.(2024)Prince, Chan, Vriza, Zhou, Sastry, Luo, Dearing, Harder, Vasudevan, and Cherukara}]{prince2024opportunities}
Michael~H Prince, Henry Chan, Aikaterini Vriza, Tao Zhou, Varuni~K Sastry, Yanqi Luo, Matthew~T Dearing, Ross~J Harder, Rama~K Vasudevan, and Mathew~J Cherukara. 2024.
\newblock Opportunities for retrieval and tool augmented large language models in scientific facilities.
\newblock \emph{npj Computational Materials}, 10(1):251.

\bibitem[{Qi et~al.(2024)Qi, Xu, Shao, Chen, Di, Cheng, Wang, and Huang}]{qi2024rora}
Jingyuan Qi, Zhiyang Xu, Rulin Shao, Yang Chen, Jin Di, Yu~Cheng, Qifan Wang, and Lifu Huang. 2024.
\newblock Rora-vlm: Robust retrieval-augmented vision language models.
\newblock \emph{arXiv preprint arXiv:2410.08876}.

\bibitem[{Qi et~al.(2023)Qi, Xu, Shen, Liu, Jin, Wang, and Huang}]{socraticqa}
Jingyuan Qi, Zhiyang Xu, Ying Shen, Minqian Liu, Di~Jin, Qifan Wang, and Lifu Huang. 2023.
\newblock \href {https://doi.org/10.18653/v1/2023.emnlp-main.255} {The art of {SOCRATIC} {QUESTIONING}: Recursive thinking with large language models}.
\newblock In \emph{Proceedings of the 2023 Conference on Empirical Methods in Natural Language Processing}, pages 4177--4199, Singapore. Association for Computational Linguistics.

\bibitem[{Reimers and Gurevych(2019)}]{reimers-gurevych-2019-sentence}
Nils Reimers and Iryna Gurevych. 2019.
\newblock \href {https://doi.org/10.18653/v1/D19-1410} {Sentence-{BERT}: Sentence embeddings using {S}iamese {BERT}-networks}.
\newblock In \emph{Proceedings of the 2019 Conference on Empirical Methods in Natural Language Processing and the 9th International Joint Conference on Natural Language Processing (EMNLP-IJCNLP)}, pages 3982--3992, Hong Kong, China. Association for Computational Linguistics.

\bibitem[{Si et~al.(2024)Si, Yang, and Hashimoto}]{si2024can}
Chenglei Si, Diyi Yang, and Tatsunori Hashimoto. 2024.
\newblock Can llms generate novel research ideas? a large-scale human study with 100+ nlp researchers.
\newblock \emph{arXiv preprint arXiv:2409.04109}.

\bibitem[{Song et~al.(2023)Song, Miret, Zhang, and Liu}]{song-etal-2023-honeybee}
Yu~Song, Santiago Miret, Huan Zhang, and Bang Liu. 2023.
\newblock \href {https://doi.org/10.18653/v1/2023.findings-emnlp.380} {{H}oney{B}ee: Progressive instruction finetuning of large language models for materials science}.
\newblock In \emph{Findings of the Association for Computational Linguistics: EMNLP 2023}, pages 5724--5739, Singapore. Association for Computational Linguistics.

\bibitem[{Sun et~al.(2024)Sun, Han, Zhao, Ma, Shen, Chen, Chen, and Yu}]{sun2024scieval}
Liangtai Sun, Yang Han, Zihan Zhao, Da~Ma, Zhennan Shen, Baocai Chen, Lu~Chen, and Kai Yu. 2024.
\newblock Scieval: A multi-level large language model evaluation benchmark for scientific research.
\newblock In \emph{Proceedings of the AAAI Conference on Artificial Intelligence}, volume~38, pages 19053--19061.

\bibitem[{Wang et~al.(2023)Wang, Fu, Du, Gao, Huang, Liu, Chandak, Liu, Van~Katwyk, Deac et~al.}]{wang2023scientific}
Hanchen Wang, Tianfan Fu, Yuanqi Du, Wenhao Gao, Kexin Huang, Ziming Liu, Payal Chandak, Shengchao Liu, Peter Van~Katwyk, Andreea Deac, et~al. 2023.
\newblock Scientific discovery in the age of artificial intelligence.
\newblock \emph{Nature}, 620(7972):47--60.

\bibitem[{Xiong et~al.(2024)Xiong, Xie, Shariatmadari, Guo, Bekiranov, and Zhang}]{xiong2024improving}
Guangzhi Xiong, Eric Xie, Amir~Hassan Shariatmadari, Sikun Guo, Stefan Bekiranov, and Aidong Zhang. 2024.
\newblock Improving scientific hypothesis generation with knowledge grounded large language models.
\newblock \emph{arXiv preprint arXiv:2411.02382}.

\bibitem[{Zhang et~al.(2024{\natexlab{a}})Zhang, Song, Hou, Miret, and Liu}]{zhang2024honeycomb}
Huan Zhang, Yu~Song, Ziyu Hou, Santiago Miret, and Bang Liu. 2024{\natexlab{a}}.
\newblock Honeycomb: A flexible llm-based agent system for materials science.
\newblock \emph{arXiv preprint arXiv:2409.00135}.

\bibitem[{Zhang et~al.(2024{\natexlab{b}})Zhang, Chen, Jin, Wang, Ji, Wang, and Han}]{zhang2024comprehensive}
Yu~Zhang, Xiusi Chen, Bowen Jin, Sheng Wang, Shuiwang Ji, Wei Wang, and Jiawei Han. 2024{\natexlab{b}}.
\newblock A comprehensive survey of scientific large language models and their applications in scientific discovery.
\newblock \emph{arXiv preprint arXiv:2406.10833}.

\bibitem[{Zheng et~al.(2014)Zheng, Lee, Weisgraber, Shusteff, DeOtte, Duoss, Kuntz, Biener, Ge, Jackson et~al.}]{zheng2014ultralight}
Xiaoyu Zheng, Howon Lee, Todd~H Weisgraber, Maxim Shusteff, Joshua DeOtte, Eric~B Duoss, Joshua~D Kuntz, Monika~M Biener, Qi~Ge, Julie~A Jackson, et~al. 2014.
\newblock Ultralight, ultrastiff mechanical metamaterials.
\newblock \emph{Science}, 344(6190):1373--1377.

\end{thebibliography}

\appendix

\newpage

\section{More Implementation Details of \system{}}

\subsection{Foundation Model for Metamaterials Science}
\label{app:foundation_model}
\subsubsection{Data Collection}
\label{app:dataclean}

To construct a high-quality dataset for metamaterials science, 
we first ask a domain expert to provide a list of keywords related to metamaterial design, retrieve and download around 5,000 research papers and books in \texttt{PDF} format from Google Scholar. We then utilize specialized toolkits, including \texttt{pdfplumber}\footnote{\url{https://github.com/jsvine/pdfplumber}}, \texttt{PyPDF2}\footnote{\url{https://github.com/py-pdf/pypdf}} and \texttt{pdfminer}\footnote{\url{https://github.com/pdfminer/pdfminer.six}}, combined with GPU-accelerated parsing\footnote{\url{https://github.com/huridocs/pdf-document-layout-analysis}} to convert the \texttt{PDF} documents to
plaintext. To ensure the quality of the extracted text, we further employ rule-based filtering and natural language processing techniques to remove conversion errors, irrelevant content, and noise from the text. Specifically, we begin by leveraging regular expressions to identify and eliminate formatting errors, symbolic noise, and other low-quality elements in the extracted text. Subsequently, natural language processing tools such as SpaCy\footnote{\url{https://spacy.io/}} and NLTK\footnote{\url{https://www.nltk.org/}} are employed for sentence segmentation, enabling the removal of incoherent or contextually irrelevant fragments. Finally, we utilize text similarity algorithms~\cite{reimers-gurevych-2019-sentence, gao-etal-2021-simcse} to detect and remove redundant paragraphs and sentences, ensuring the uniqueness and relevance of processed text.

Finally, we categorize the cleaned text corpus into two subsets: (1) A \textbf{domain-specific} subset that includes academic papers and textbooks specifically focused on metamaterials science. Accounting for 80\% of the whole training corpus, this subset provides specialized knowledge and ensures a deep focus on the target domain. (2) A \textbf{general-domain} subset that contains scientific papers about other topics in materials science and accounts for 20\% of the training corpus. This subset offers foundational knowledge about the board materials science field, complementing the domain-specific insights. Table~\ref{tab:data_statistics} shows detailed statistics of the dataset.

\begin{table}[h]
\resizebox{0.5\textwidth}{!}
{
\begin{tabular}{l|c|c}
\toprule
\textbf{Category}         & \textbf{Count}     & \textbf{Details} \\
\midrule
Total \texttt{pdf} Files      & 5611               & {Approximately 480 Books}      \\
\midrule
Domain-Specific Papers    & 4470               & {MetaMaterial Science} \\
General-Domain Papers     & 1141               & {General Material Science} \\
\midrule
Converted \texttt{txt} Files       & 4711               & After conversion \\
Words in Cleaned Data     & 42,488,911         & (42M)           \\
Tokens in Cleaned Data    & 74,814,532         & (74M)           \\
\bottomrule
\end{tabular}
}
\caption{Dataset Statistics for the Foundation Model}
\label{tab:data_statistics}
\end{table}

\subsubsection{Model Training and Evaluation}
\label{app:modeltrain}
The model was trained using a learning rate of \(2 \times 10^{-4}\) and the \texttt{Adam} optimizer~\cite{kingma2014adam} with $\epsilon=1 \times 10^{-8}$. A cosine learning rate scheduler was employed, with a warmup ratio of \(0.1\). For fine-tuning, we utilized LoRA with \texttt{lora\_alpha}=8 and a dropout rate of 0.05. The training was conducted on \(4 \times\) RTX A6000 GPUs. To ensure optimal performance, models were trained across different numbers of epochs, and the version trained for 5 epochs was selected as the final iteration.

\begin{figure*}[ht!]
    \centering
    \includegraphics[width=0.95\linewidth, trim={0 0 0 0},clip ]{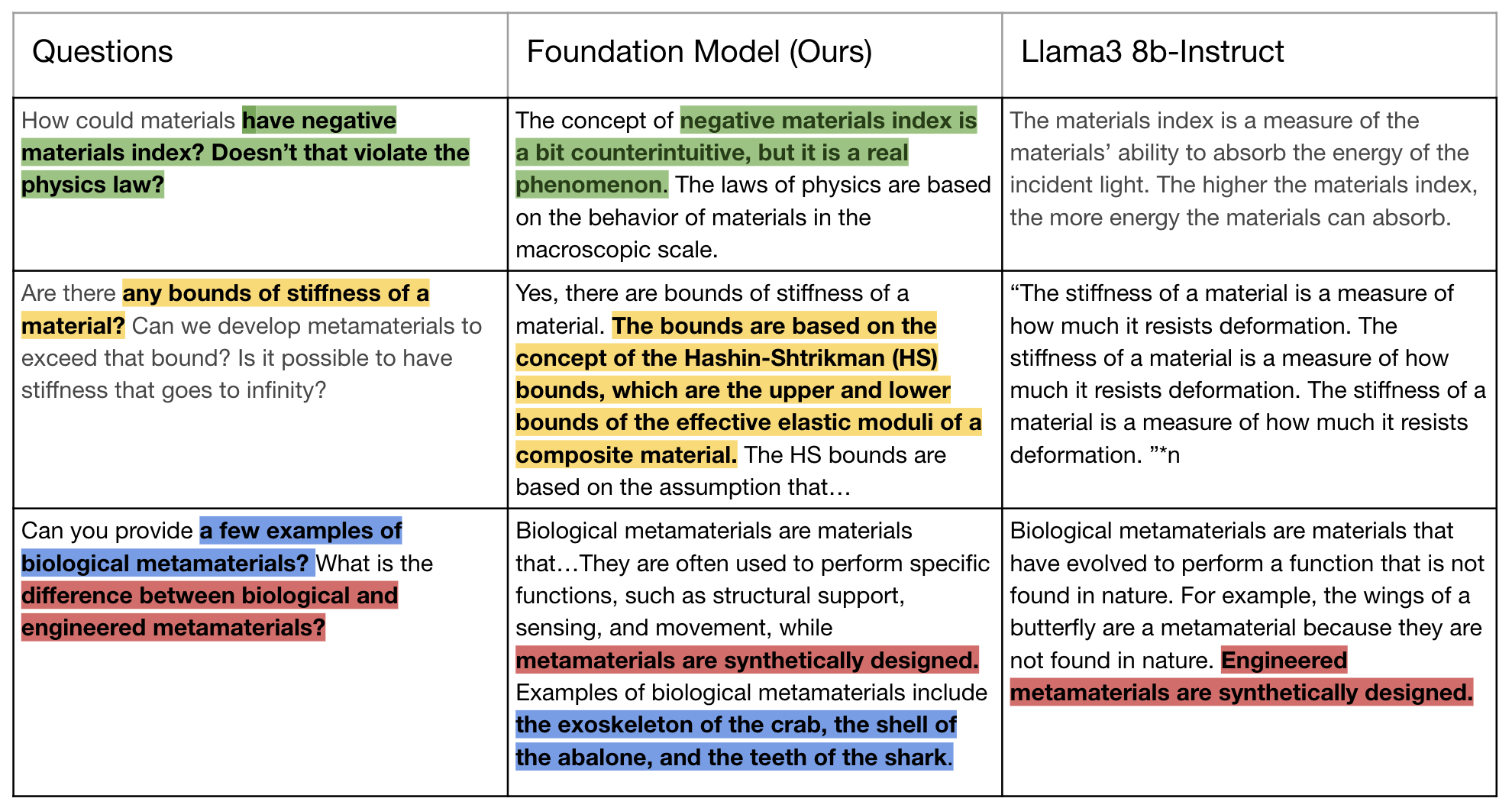}
	\caption{Comparison of answers provided by the Foundation Model (Ours) and LLaMA3-8B-Instruct to questions about material science.}
    \vspace{-5mm}
\label{fig:qa_foundation}
\end{figure*}

\paragraph{Comparison of Foundation model and Llama 3-8B}
As shown in Figure \ref{fig:qa_foundation}, the comparison highlights the differences in the responses generated by the Foundation Model (Ours) and the baseline LLaMA3-8B-Instruct model for material science-related questions. The Foundation Model exhibits a more detailed and contextually grounded understanding, integrating advanced scientific concepts such as the negative material index and the Hashin-Shtrikman bounds. Additionally, it provides clear examples of biological metamaterials, including the exoskeleton of the crab and the shell of the abalone, demonstrating a deeper domain-specific reasoning capability. 

In contrast, the responses from the LLaMA3-8B-Instruct model are either overly generalized, repetitive, or lack sufficient scientific detail to address the questions effectively. This qualitative comparison underscores the improved performance of the Foundation Model in tasks requiring domain knowledge and contextual comprehension.

\begin{figure*}[ht!]
    \centering
    \includegraphics[width=0.85\linewidth, trim={0 0 0 0},clip ]{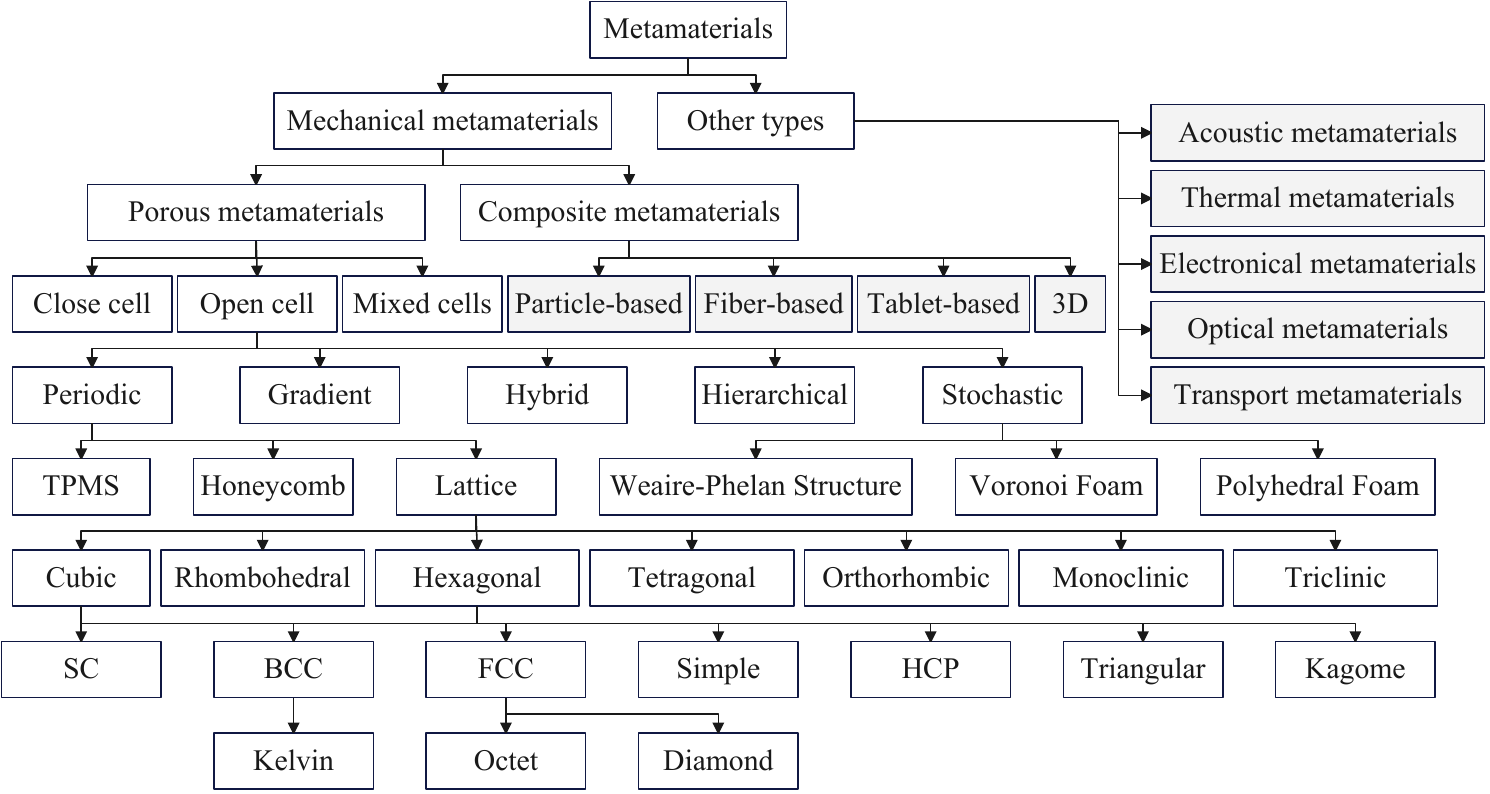}
	\caption{Illustration of our curated taxonomy tree in mechanical metamaterials used in internal inductive bias extraction. }
\label{fig:taxonomy1}
\end{figure*}

\begin{figure*}[ht!]
    \centering
    \includegraphics[width=0.95\linewidth, trim={0 0 0 0},clip ]{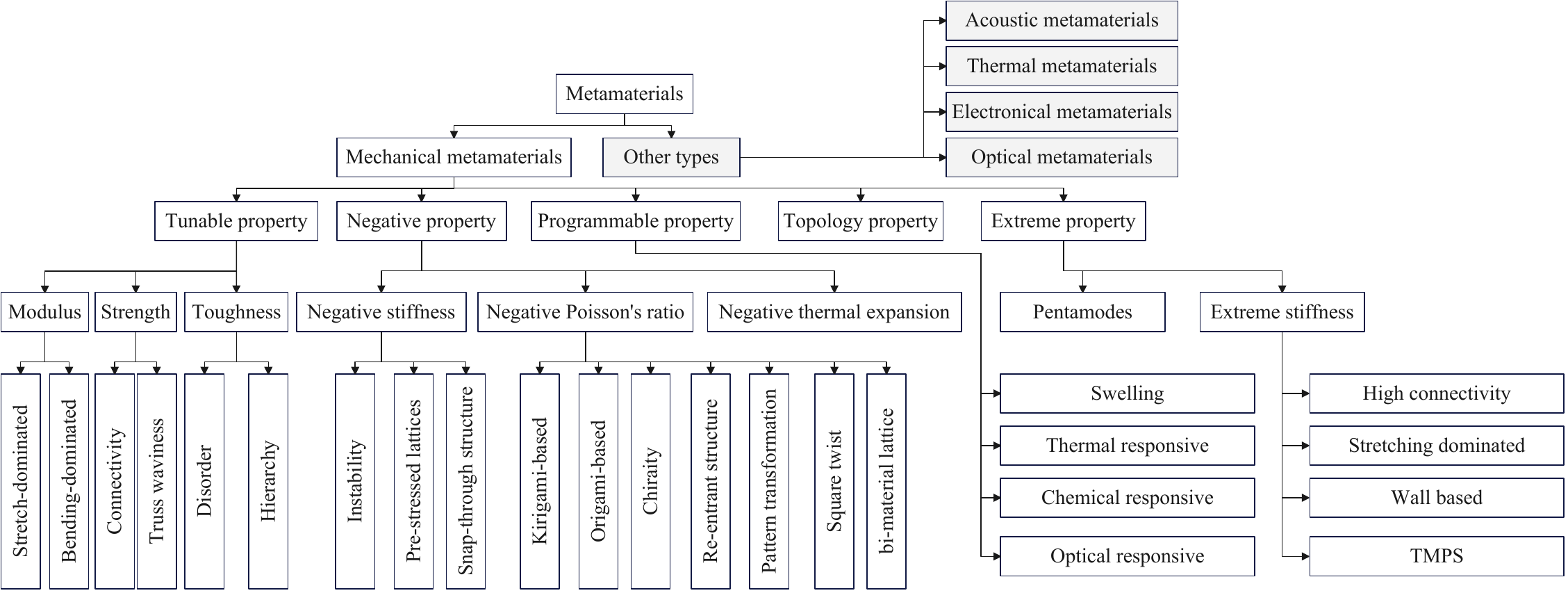}
	\caption{Illustration of our curated taxonomy tree in mechanical metamaterials used in internal inductive bias extraction. }
\label{fig:taxonomy2}
\end{figure*}

\subsection{Hypothesis Generation}
\label{app:hypo_gen}

\paragraph{More details on Inductive Biases Extraction} We let a domain expert in metamaterial science domain curate a metamaterial taxonomy for eliciting internal relevant knowledge from LLMs as shown in Figure~\ref{fig:taxonomy1} and Figure~\ref{fig:taxonomy2}.

\subsection{3D Structure Synthesis}
\label{app:3d_syn}

\paragraph{Training Details for 3D Structure Generation} To train our 3D structure synthesis models, we extend the Modulus~\cite{lumpe2021exploring} dataset by applying scaling and rotation transformations to its lattice topologies, paired with adjusted lattice properties to ensure the dataset captures a broader range of variations. During training, the diffusion block receives 3D coordinates mixed with Gaussian noise as input, along with the corresponding lattice properties as conditioning information. The model predicts the clean 3D coordinates in a single step, and the loss is computed based on the difference between the predicted and ground truth coordinates. Simultaneously, the edge prediction block is trained by comparing its predictions with ground truth vertex connections. This joint training process enables the model to generate accurate and structurally coherent 3D lattice designs.

\paragraph{Training Details for 3D Structure Refinement} 
The major difficulty in training the refinement model is the limited availability of paired data showing both unrefined and properly refined lattice structures. While researchers typically publish only their optimized final structures, they often omit documenting the optimization process itself. To overcome this obstacle, we synthesize training data based on the Unit Cell Catalog~\cite{lumpe2021exploring}. The diffusion-generated lattices exhibited two primary failure patterns: inaccurate node coordinates and incorrect node/edge configurations (either missing or excessive). To mimic these failure patterns, we corrupt the perfect lattice by randomly remove or add nodes and edges, and inject Gaussian noise into the node coordinates. Hereafter, we supervised finetune the pretrained model (Llama-3.1-8B-Instruct \citep{dubey2024llama}) by fitting the lattice structure information into the prompt and response template. The model gets the corrupted lattice in the prompt and is trained to include the perfect lattice in the response.

\section{More Experiment Details and Results}
\label{app:experiment}

\subsection{Case Studies}
\label{app:case_study}
\begin{figure*}[ht!]
    \centering
    \includegraphics[width=\linewidth, trim={0 0 0 0},clip ]{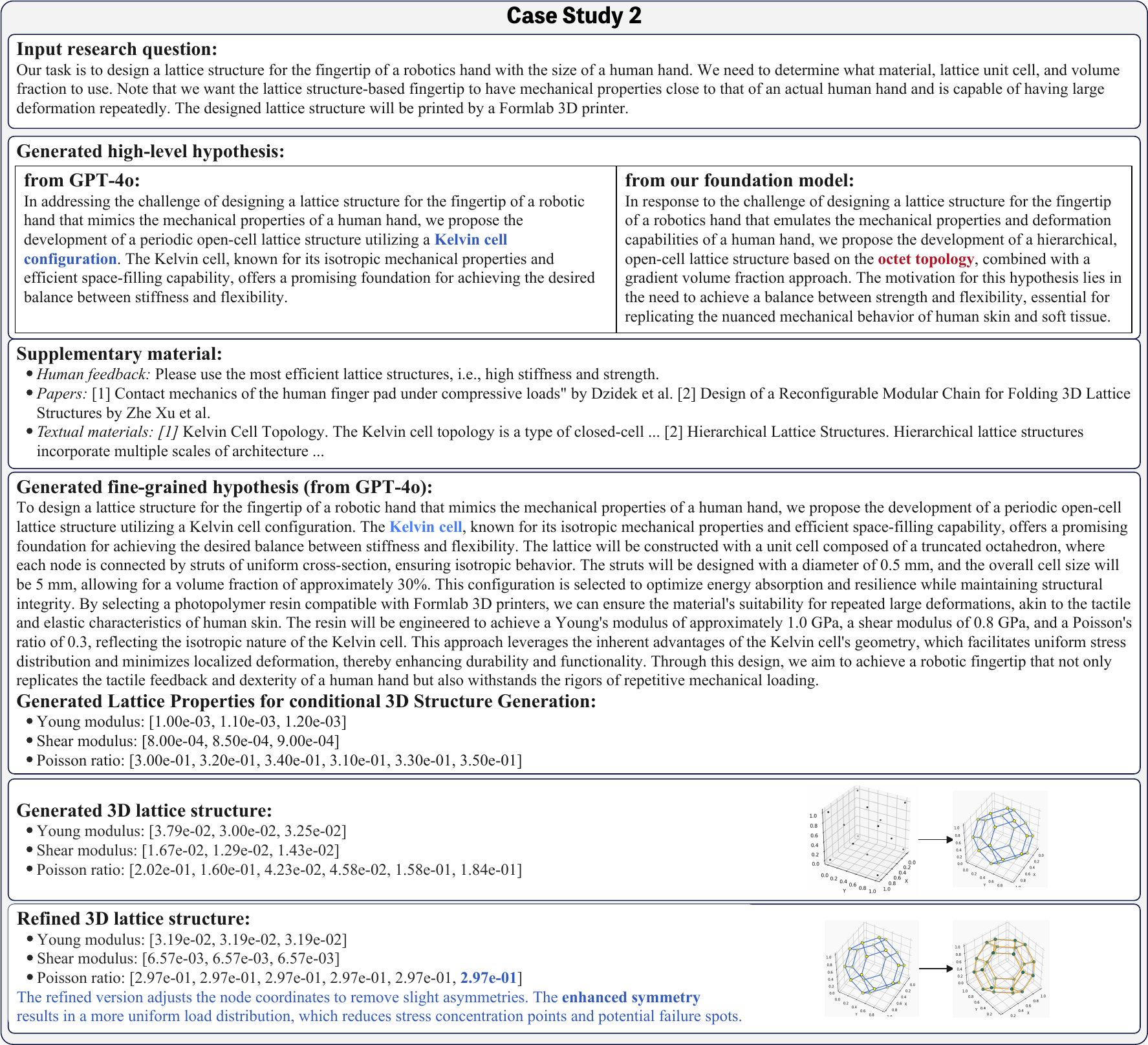}
	\caption{\textbf{Case study} of \textit{designing a lattice structure for the fingertip of a robotics hand} using \system{}. }
\label{casestudy2}
\end{figure*}

We present additional case studies in Figure~\ref{casestudy2}. 
From this case study focusing on designing a lattice structure for a robotic fingertip that mimics human hand properties, we have the following observations.
\textbf{First,} our system effectively incorporates domain expertise through human feedback and supplementary materials. Though our fine-tuned foudation model proposed a more sophisticated hierarchical octet topology with gradient volume fraction, when provided with expert feedback emphasizing high stiffness and strength requirements, along with relevant papers on human finger pad mechanics and 3D lattice structures, our systems converged on the Kelvin cell configuration focusing primarily on isotropy. The supplementary materials also helped establish reasonable ranges for mechanical properties in the fine-grained hypothesis, particularly in matching human tissue characteristics. \textbf{Second,} we demonstrate that our refinement model can successfully optimize the generated lattice structure for improved mechanical symmetry. The initial structure showed variations in Young's modulus (ranging from 3.00e-02 to 3.79e-02) and significant anisotropy in Poisson ratios (ranging from 4.23e-02 to 2.02e-01). Through our refinement process, these properties were harmonized to achieve perfect cubic symmetry with consistent values (Young's modulus of 3.19e-02 and Poisson ratio of 2.97e-01).

\subsection{More Details and Results of Human Evaluation for Hypothesis Generation}
\label{app:hypogen_eval}

\paragraph{Human Evaluation}
We evaluate the generated hypotheses with two aspects, i.e., \textit{novelty} and \textit{feasibility}. 
We ask three senior researchers in the metamaterial domain as the reviewers to score the system-generated hypotheses. Each reviewer gives the scores independently.
The evaluation criteria are defined as follows:

\textbf{Novelty:} Whether the idea is creative and different from existing works on the topic, and brings fresh insights.
\begin{itemize}
    \item \textbf{Score 0 (Low Novelty): }  The idea closely resembles many existing works with minimal differentiation.
    \item \textbf{Score 1 (Medium Novelty):} The idea shows some differences from existing works but has obvious overlaps with prior concepts. 
    \item \textbf{Score 2 (High Novelty):}  The idea introduces major differences, offering distinct insights or approaches absent in existing works.
\end{itemize}

\textbf{Feasibility:} How feasible is it to implement and execute this idea as a research work.
\begin{itemize}
    \item \textbf{Score 0 (Low Feasibility): } The idea is highly impractical or requires resources/techniques currently unavailable.
    \item \textbf{Score 1 (Medium Feasibility): } The idea is somewhat practical but faces notable challenges in execution or resource availability.
    \item \textbf{Score 2 (High Feasibility): } The idea is practical and can be implemented with available resources and techniques.
\end{itemize}

The complete list of the examples used in human evaluation can be found in Appendix ~\ref{sec:eval_example}.

\section{Prompts used in \system{}}
\begin{figure*}[!tbp]
    \centering
    \includegraphics[width=\linewidth, trim={0 0 0 0},clip ]{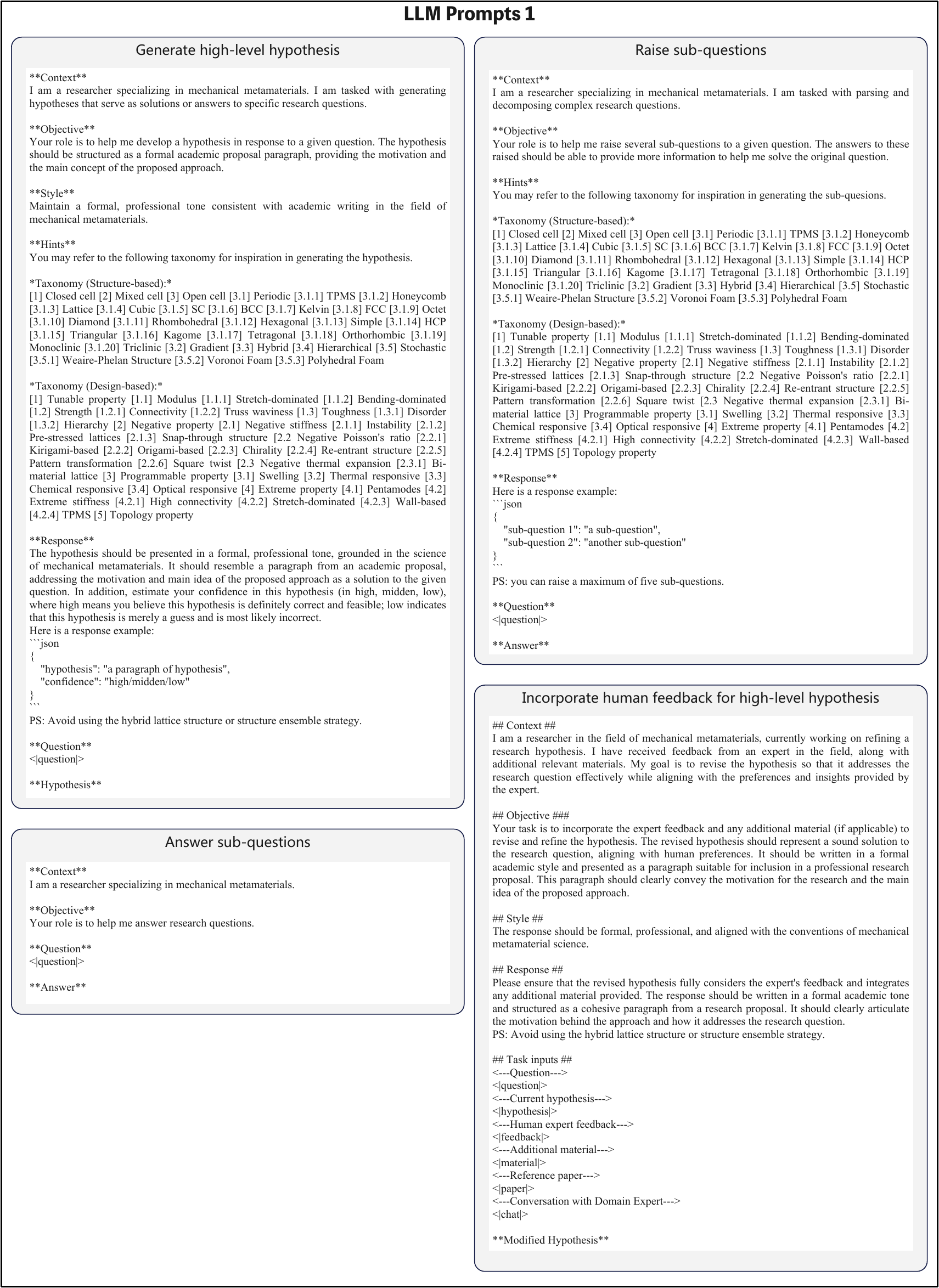}
	\caption{Prompts used in \system{} (part 1)}
\label{prompts1}
\end{figure*}

\begin{figure*}[!tbp]
    \centering
    \includegraphics[width=\linewidth, trim={0 0 0 0},clip ]{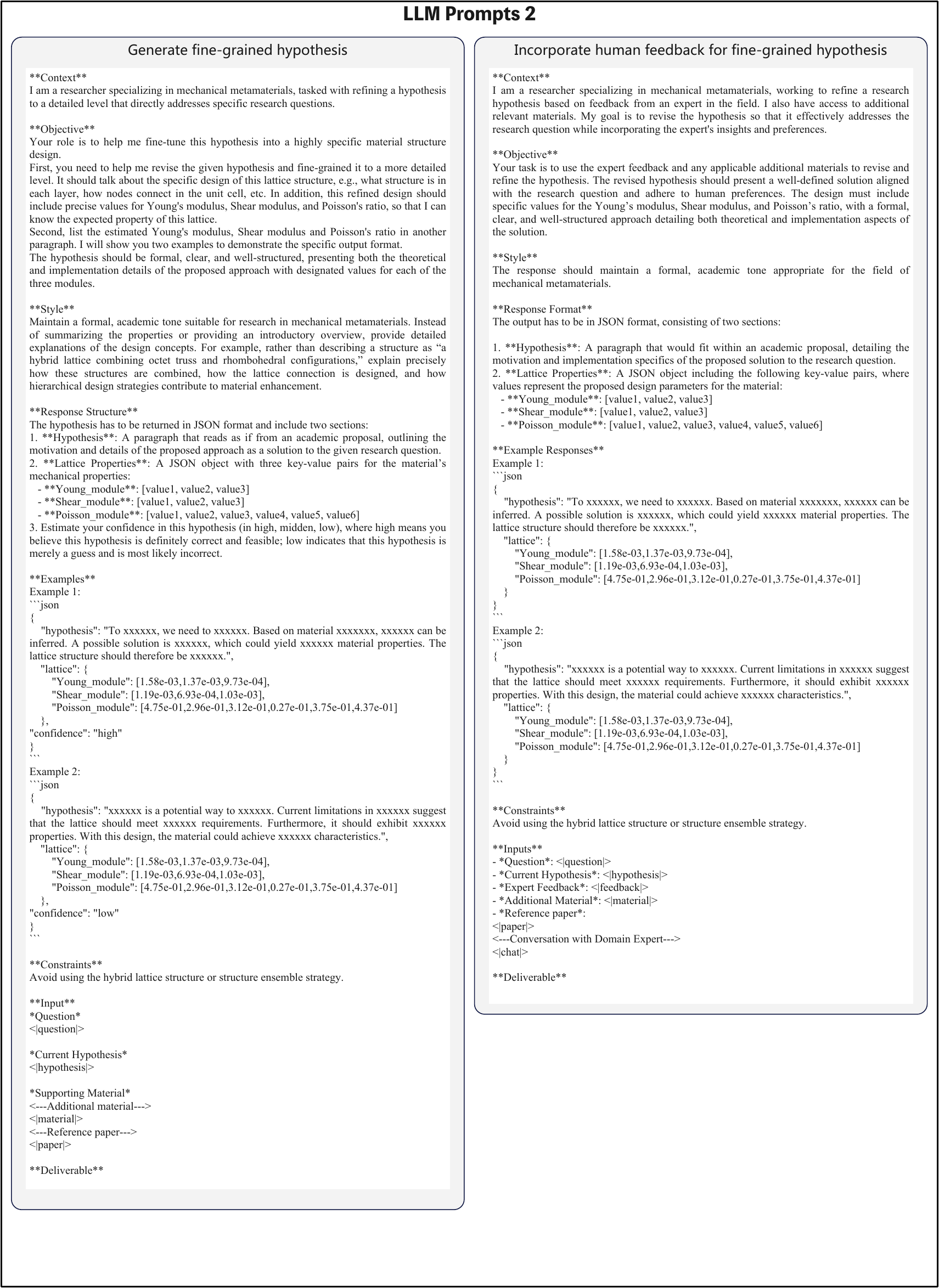}
	\caption{Prompts used in \system{} (part 2)}
\label{prompts2}
\end{figure*}

\begin{figure*}[!tbp]
    \centering
    \includegraphics[width=\linewidth, trim={0 0 0 0},clip ]{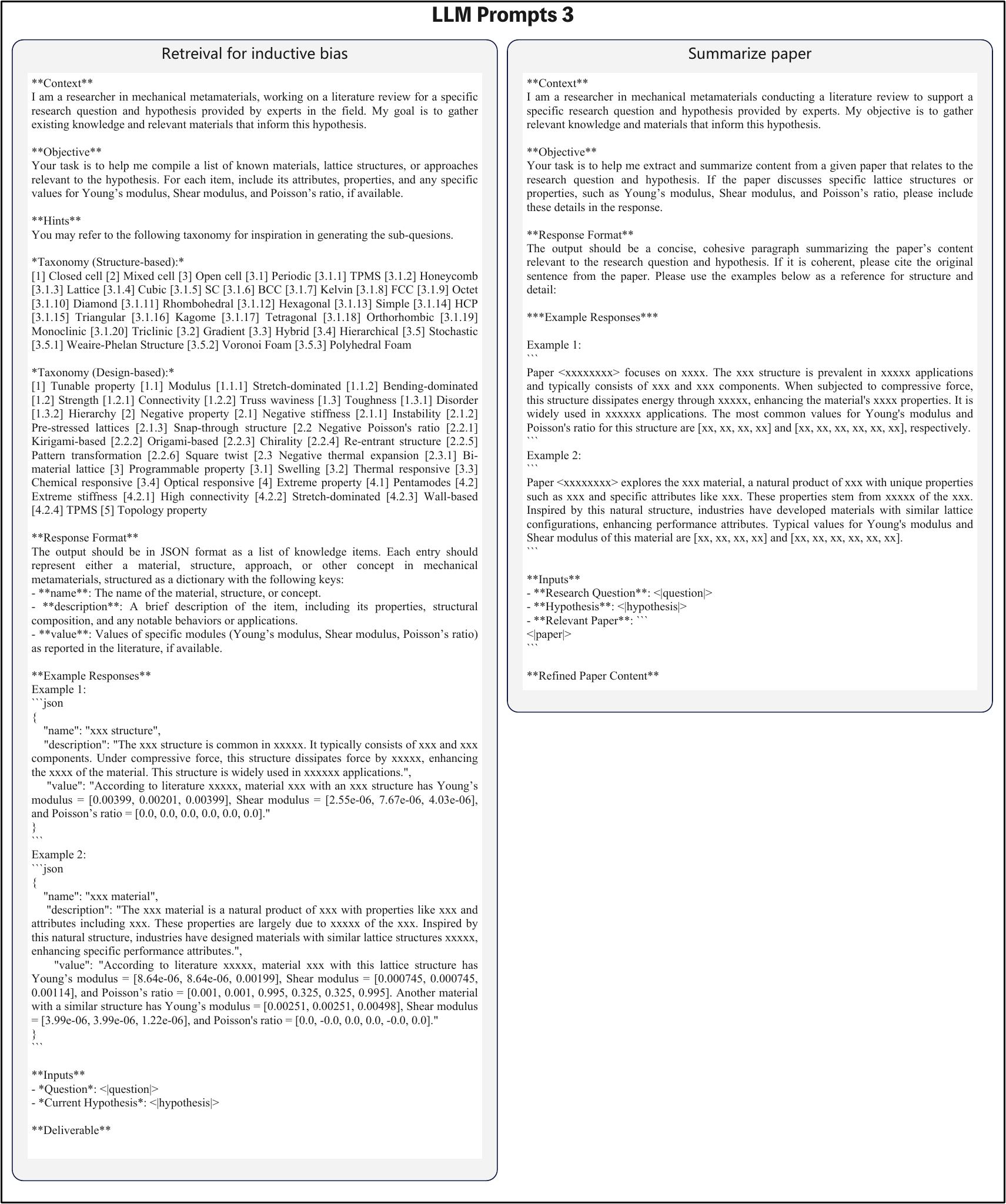}
	\caption{Prompts used in \system{} (part 3)}
\label{prompts3}
\end{figure*}

We offer all prompts we used in the \system{} in Figure~\ref{prompts1},~\ref{prompts2}, and~\ref{prompts3}.

\onecolumn
\section{Examples for human evaluation}
\label{sec:eval_example}
\begin{itemize}
    \item \textbf{Example 1}
    \begin{itemize}
        \item \textbf{Topic:} novel lattice metamaterials 
        \item \textbf{Question:} Can you propose a novel scientific hypothesis on designing lattice metamaterials that are drastically different from existings concepts? The concept should have solid reasoning, and the designs generated based on the hypothesis should be testible and can be easily manufactured.
        \item \textbf{GPT-4o:}
        \begin{itemize}
            \item \textbf{Hypothesis:} In the pursuit of designing lattice metamaterials that diverge significantly from existing concepts, we propose the development of a novel class of metamaterials based on a hierarchical, multi-scale approach that integrates periodic and stochastic elements. This innovative design paradigm leverages the inherent advantages of both periodic structures, such as TPMS and BCC lattices, and stochastic configurations like Voronoi foams, to achieve unprecedented mechanical properties. The motivation for this approach stems from the need to overcome the limitations of traditional lattice designs, which often exhibit trade-offs between strength, toughness, and manufacturability. By embedding stochastic elements within a periodic framework, we hypothesize that it is possible to enhance the energy absorption capacity and damage tolerance of the metamaterial while maintaining structural integrity and ease of fabrication. This dual-scale design can be realized through advanced additive manufacturing techniques, which allow for precise control over the micro-architecture. The proposed metamaterials are expected to exhibit tunable mechanical properties, such as modulus and toughness, by adjusting the scale and distribution of the stochastic elements within the periodic matrix. This hypothesis not only provides a solid foundation for the development of next-generation metamaterials but also opens new avenues for experimental validation and practical applications in fields requiring lightweight, high-performance materials.
            \item \textbf{Reviewer Comment:} Lack of novelty. The combination of periodic and stochastic structures in a dual-scale configuration is not explained.
            \item \textbf{Novelty:} [0,1,1]
            \item \textbf{Feasibility:} [1,2,2]
        \end{itemize}
        \item \textbf{Metascientist without human:}
        \begin{itemize}
            \item \textbf{Hypothesis:} To address the challenge of designing lattice metamaterials that are drastically different from existing concepts, we propose a novel approach that leverages a dual-layered lattice structure with distinct mechanical properties. The design consists of a primary layer featuring a hexagonal close-packed (HCP) configuration, known for its high packing density and isotropic mechanical properties, and a secondary layer composed of a diamond cubic lattice, which provides excellent stiffness and low density. The nodes in the HCP layer are connected through cylindrical struts with a diameter of 0.5 mm, while the diamond cubic layer utilizes struts with a diameter of 0.3 mm. The integration of these two layers is achieved through a hierarchical interlocking mechanism, where the nodes of the diamond cubic lattice are embedded within the interstitial spaces of the HCP layer, ensuring seamless load transfer and enhanced structural integrity. This design is expected to exhibit a Young's modulus of approximately 1.1e-3, a shear modulus of 8.5e-4, and a Poisson's ratio of 0.32, providing a balance between stiffness, strength, and manufacturability. The proposed metamaterial can be fabricated using advanced additive manufacturing techniques, allowing for precise control over the micro-architecture and enabling the exploration of tunable mechanical properties by adjusting the relative thickness and orientation of the layers. This hypothesis not only offers a new direction for the development of next-generation metamaterials but also provides a testable and manufacturable solution for applications requiring lightweight, high-performance materials.
            \item \textbf{Reviewer Comment:} Lack of novelty. How the two layers are connected at the interstitial of HCP layer is not clear.
            \item \textbf{Novelty:} [0,1,0]
            \item \textbf{Feasibility:} [1,1,2]
        \end{itemize}
        \item \textbf{Metascientist with human:}
        \begin{itemize}
            \item \textbf{Hypothesis:} In response to the challenge of designing lattice metamaterials that diverge significantly from existing paradigms, we propose a novel hypothesis centered on the integration of phase-change principles at the lattice structure level to enable dynamic adaptability. This approach is motivated by the need for metamaterials that can alter their mechanical properties in real-time to meet diverse application requirements, such as in aerospace components that adapt to varying flight conditions or biomedical devices that respond to physiological changes. By leveraging phase-change mechanisms within the lattice architecture itself, rather than at the material level, we can achieve reversible transformations in the structural configuration in response to external stimuli like temperature or electrical inputs. This structural phase-change capability allows for the modulation of mechanical properties, including stiffness, damping, and energy absorption, providing a versatile platform for tailored performance. The hypothesis posits that by strategically designing lattice geometries that can undergo controlled reconfiguration through localized heating or electrical activation, we can achieve a metamaterial system capable of transitioning between multiple functional states, thereby offering unprecedented adaptability. This concept not only aligns with current research trends but also opens new avenues for exploration in the field of mechanical metamaterials. The proposed designs are amenable to fabrication using contemporary additive manufacturing techniques, ensuring their feasibility and facilitating empirical validation.
            \item \textbf{Reviewer Comment:} While the idea is still not new, using phase-change principles is interesting. 
            \item \textbf{Novelty:} [1,2,2]
            \item \textbf{Feasibility:} [2,1,1]
        \end{itemize}
    \end{itemize}

    \item \textbf{Example 2}
    \begin{itemize}
        \item \textbf{Topic:} ultra efficient structural materails
        \item \textbf{Question:} Can you generate some really novel scientific hypotheses in how to improve the structural efficiency of lattice materials, which can be validated or falsified relatively easily by experiments, but are also quite interesting and important?
        \item \textbf{GPT-4o:}
        \begin{itemize}
            \item \textbf{Hypothesis:} In response to the question of improving the structural efficiency of lattice materials, we propose the hypothesis that incorporating a hierarchical design approach within a periodic open-cell lattice structure, specifically utilizing a combination of TPMS (Triply Periodic Minimal Surfaces) and octet truss configurations, can significantly enhance the mechanical performance of these materials. The motivation for this hypothesis stems from the inherent advantages of hierarchical structures, which are known to provide superior strength-to-weight ratios and energy absorption capabilities due to their multi-scale architecture. By integrating TPMS, known for its minimal surface area and high stiffness, with the octet truss, recognized for its stretch-dominated behavior, we anticipate achieving a synergistic effect that optimizes both stiffness and strength. This approach not only leverages the geometric complexity of TPMS to distribute stress uniformly but also utilizes the efficient load-bearing capacity of the octet truss. The proposed hypothesis can be validated through experimental fabrication and mechanical testing of lattice prototypes, assessing parameters such as compressive strength, stiffness, and energy absorption. This investigation holds significant potential for advancing the design of lightweight, high-performance materials in applications ranging from aerospace to biomedical engineering.
            \item \textbf{Reviewer Comment:} Hierachy is a well known design stragety. Hierarchy does not necessarily improve the mechanical efficiency.
            \item \textbf{Novelty:} [0,1,1]
            \item \textbf{Feasibility:} [2,2,2]
        \end{itemize}
        \item \textbf{Metascientist without human:}
        \begin{itemize}
            \item \textbf{Hypothesis:} To improve the structural efficiency of lattice materials, we propose a novel hypothesis that involves the design of a multi-layered lattice structure with a specific focus on a gradient-based transition between two distinct configurations: a gyroid-based TPMS and a diamond-based TPMS. The design will consist of three layers, each with a unique configuration to optimize mechanical properties. The top layer will utilize a gyroid TPMS structure, characterized by its high stiffness and minimal surface area, to provide initial load distribution and stress uniformity. The middle layer will transition into a diamond TPMS structure, which offers enhanced energy absorption and strength due to its interconnected nodal design. The bottom layer will revert to a gyroid TPMS, ensuring a balanced load transfer and structural integrity. Nodes within each unit cell will be connected using a continuous curvature approach to minimize stress concentrations and enhance load-bearing capacity. This hierarchical design strategy aims to achieve a synergistic effect, optimizing both stiffness and strength while maintaining a lightweight structure. The proposed hypothesis can be validated through experimental fabrication and mechanical testing, focusing on parameters such as compressive strength, stiffness, and energy absorption. This approach holds significant potential for advancing the design of lightweight, high-performance materials in applications ranging from aerospace to biomedical engineering.
            \item \textbf{Reviewer Comment:} Lack of novelity, but the proposed structure design is very specific.
            \item \textbf{Novelty:} [0,1,0]
            \item \textbf{Feasibility:} [1,2,2]
        \end{itemize}
        \item \textbf{Metascientist with human:}
        \begin{itemize}
            \item \textbf{Hypothesis:} In response to the question of improving the structural efficiency of lattice materials, this revised hypothesis proposes the development of a bio-inspired, hierarchical lattice structure that leverages the synergistic enhancement of mechanical properties. By exploring the intricate microstructural organization found in natural materials such as the nacreous layers of mollusk shells and the micro-architectures within coral skeletons, which exhibit remarkable mechanical efficiency through multi-scale architectures, this approach aims to integrate nano-sized features for local stiffness and strength, transitioning to larger structural elements for overall stability and load distribution. The hypothesis suggests that material gradation, akin to natural transitions between hard and soft regions, will promote stress distribution and energy absorption, thereby enhancing the structural efficiency of lattice materials. This innovative pathway, grounded in the meticulous analysis of diverse biological precedents, promises to significantly advance the field of mechanical metamaterials through experimental validation and computational modeling.
            \item \textbf{Reviewer Comment:} Bioinspired strageties are quite common in research, but it is a highly feasible approach.
            \item \textbf{Novelty:} [1,2,1]
            \item \textbf{Feasibility:} [2,1,1]
        \end{itemize}
    \end{itemize}

    \item \textbf{Example 3}
    \begin{itemize}
        \item \textbf{Topic:} improving damage tolerance
        \item \textbf{Question:} One critical limitation to cellular ceramic materials are their brittleness and low damage tolerance. Can you propose some novel scientific hypotheses in how to improve the damage tolerance of lattice materials, which can be validated and are feasible to produce based on current manufacturing approaches?
        \item \textbf{GPT-4o:}
        \begin{itemize}
            \item \textbf{Hypothesis:} To address the inherent brittleness and low damage tolerance of cellular ceramic materials, we propose the development of a hierarchical lattice structure that integrates a combination of periodic and stochastic design elements. This approach leverages the mechanical advantages of both periodic and stochastic structures to enhance energy absorption and crack deflection capabilities. Specifically, we hypothesize that embedding a periodic lattice framework, such as a Kelvin or BCC structure, with stochastic elements like Voronoi foam, will create a synergistic effect that improves the overall toughness of the material. The periodic lattice provides a predictable load distribution and structural integrity, while the stochastic elements introduce randomness that can disrupt crack propagation pathways, thereby enhancing damage tolerance. This hybrid design can be feasibly manufactured using advanced additive manufacturing techniques, such as selective laser sintering or stereolithography, which allow for precise control over complex geometries at multiple scales. By optimizing the hierarchical arrangement and material composition, this novel lattice structure is expected to exhibit superior mechanical performance, making it a promising candidate for applications requiring high damage tolerance in ceramic materials.
            \item \textbf{Reviewer Comment:} Structural radomness is a method proposed reletively recently, but it's not new. The usage of Kelvin or BCC lattices cannot grarantee high stiffness and strength, which impedes the feasibility of the hypothesis.
            \item \textbf{Novelty:} [1,1,1]
            \item \textbf{Feasibility:} [1,2,2]
        \end{itemize}
        \item \textbf{Metascientist without human:}
        \begin{itemize}
            \item \textbf{Hypothesis:} To improve the damage tolerance of cellular ceramic materials, we propose the design of a novel lattice structure that employs a multi-layered approach, integrating a periodic Kelvin lattice at the core with a surrounding layer of stochastic Voronoi foam. The core Kelvin lattice, composed of truncated octahedra, ensures efficient load distribution and structural integrity, while the outer Voronoi foam layer introduces randomness to disrupt crack propagation pathways. This design leverages the mechanical stability of the Kelvin lattice, with its high strength-to-weight ratio, and the energy absorption capabilities of the stochastic layer. The nodes within the unit cell of the Kelvin lattice are connected via struts that are optimized for axial and bending loads, ensuring a balance between stiffness and flexibility. The hierarchical arrangement is further enhanced by varying the strut thickness and material composition across layers, allowing for tailored mechanical properties. This structure is expected to exhibit a Young's modulus of approximately 1.1e-3, a shear modulus of 8.5e-4, and a Poisson's ratio of 0.32, making it a promising candidate for applications requiring high damage tolerance. The proposed lattice can be feasibly manufactured using advanced additive manufacturing techniques, such as selective laser sintering, which allow for precise control over complex geometries and material distribution.
            \item \textbf{Reviewer Comment:} Introducing structural radomness to enhance toughness is a method proposed reletively recently. The usage of Kelvin or BCC lattices cannot grarantee high stiffness and strength, which impedes the feasibility of the hypothesis. Additionally, the values of Young's modulus of approximately 1.1e-3, a shear modulus of 8.5e-4, and Poisson's ratio of 0.32 lack reasoning.
            \item \textbf{Novelty:} [1,1,1]
            \item \textbf{Feasibility:} [0,1,2]
        \end{itemize}
        \item \textbf{Metascientist with human:}
        \begin{itemize}
            \item \textbf{Hypothesis:} To enhance the damage tolerance of ceramic lattice materials, this research proposes the development of a multi-step interlocking interface inspired by the sacrificial bonds found in bone. The key design principle involves creating a hierarchical interlocking mechanism within the lattice structure, consisting of primary, secondary, and tertiary interlocks. These interlocks are strategically designed to engage sequentially under increasing loads, providing multiple levels of energy dissipation and crack deflection. By mimicking the multi-level sacrificial bonding mechanism of bone, this approach aims to significantly improve the fracture toughness and damage tolerance of ceramic materials. The hypothesis will be validated through computational modeling and experimental testing, focusing on optimizing the geometry and material distribution to achieve maximum toughness without compromising structural integrity.
            \item \textbf{Reviewer Comment:} multi-step interlocking interface inspired by the sacrificial bonds of bone is interesting.
            \item \textbf{Novelty:} [2,2,2]
            \item \textbf{Feasibility:} [2,1,2]
        \end{itemize}
    \end{itemize}

    \item \textbf{Example 4}
    \begin{itemize}
        \item \textbf{Topic:} new auxetic metamaterials
        \item \textbf{Question:} How can we innovate a new auxetic metamaterials that have negative Possion's ratio yet keep a high relative modulus? Please provide detailed design idea and highlight its potential advantages compared to other existing auxetic metamaterials.
        \item \textbf{GPT-4o:}
        \begin{itemize}
            \item \textbf{Hypothesis:} In response to the challenge of developing a novel auxetic metamaterial that exhibits a negative Poisson's ratio while maintaining a high relative modulus, we propose the design and fabrication of a hierarchical re-entrant structure integrated with a stretch-dominated lattice framework. The motivation for this approach stems from the need to overcome the traditional trade-off between auxetic behavior and mechanical stiffness, which often limits the practical applications of auxetic materials. By employing a hierarchical design, we can leverage the benefits of multiple length scales to enhance mechanical properties. The re-entrant structure, known for its auxetic characteristics, will be strategically embedded within a stretch-dominated lattice, which is inherently capable of supporting high loads due to its efficient load distribution and connectivity. This combination is expected to synergistically enhance the overall mechanical performance, providing a metamaterial that not only exhibits a negative Poisson's ratio but also achieves a high relative modulus. Compared to existing auxetic metamaterials, this design offers the potential advantage of improved mechanical robustness and versatility, making it suitable for applications requiring both flexibility and strength, such as in advanced protective gear, biomedical devices, and adaptive architectural structures. The proposed metamaterial could thus represent a significant advancement in the field, offering a new paradigm for the design of multifunctional materials.
            \item \textbf{Reviewer Comment:} Lack of novelity. Re-entrant structure and a stretch-dominated lattice framework does not simply add together. How to achive this is not explained.
            \item \textbf{Novelty:} [0,1,1]
            \item \textbf{Feasibility:} [0,2,2]
        \end{itemize}
        \item \textbf{Metascientist without human:}
        \begin{itemize}
            \item \textbf{Hypothesis:} To innovate a new auxetic metamaterial that exhibits a negative Poisson's ratio while maintaining a high relative modulus, we propose the design of a hierarchical re-entrant honeycomb structure integrated with a stretch-dominated lattice framework. This design aims to address the traditional trade-off between auxetic behavior and mechanical stiffness. The re-entrant honeycomb structure, characterized by its inverted angles, will be strategically embedded within a stretch-dominated lattice. The unit cell will consist of a central re-entrant honeycomb core surrounded by a network of axially aligned struts forming a stretch-dominated lattice. Each layer of the lattice will alternate between re-entrant and stretch-dominated configurations, with nodes connecting at the vertices of the re-entrant angles and the endpoints of the struts. This configuration allows for efficient load distribution and enhanced mechanical properties. The hierarchical design leverages multiple length scales to optimize the material's mechanical performance, achieving a negative Poisson's ratio while maintaining a high relative modulus. Compared to existing auxetic metamaterials, this design offers improved mechanical robustness and versatility, making it suitable for applications requiring both flexibility and strength, such as advanced protective gear, biomedical devices, and adaptive architectural structures.
            \item \textbf{Reviewer Comment:} Lack of novelity.The proposed structure arrangement is very detailed but does not make sense.
            \item \textbf{Novelty:} [0,1,2]
            \item \textbf{Feasibility:} [0,2,2]
        \end{itemize}
        \item \textbf{Metascientist with human:}
        \begin{itemize}
            \item \textbf{Hypothesis:} In response to the challenge of developing auxetic metamaterials that exhibit a negative Poisson's ratio while maintaining a high relative modulus, we propose a novel approach that eschews the traditional reliance on reentrant joints and folding mechanisms, which are often associated with reduced mechanical efficiency and increased manufacturing complexity. Instead, our design leverages a plate-based structure that integrates strategic perforated patterns with material anisotropy and gradient material properties. By utilizing tessellated perforations, such as hexagonal or triangular grids, and capitalizing on the directional properties of advanced composite materials like fiber-reinforced polymers, the structure is engineered to expand laterally under tensile loads, thereby achieving the desired auxetic behavior. The incorporation of a gradient in material properties further optimizes stress distribution, enhancing the auxetic response while maintaining structural integrity. This innovative design not only addresses the traditional trade-off between auxeticity and stiffness but also ensures high mechanical efficiency, scalability, and manufacturability using conventional techniques. Moreover, it offers significant versatility, allowing for customization to meet specific application requirements across diverse fields such as aerospace, automotive, and consumer products. By focusing on these advanced design principles, our approach presents a promising pathway for advancing the capabilities and applications of auxetic metamaterials.
            \item \textbf{Reviewer Comment:} The hypotheis address the critial constrain that reentrant and folding are often necessary for achieving auxetic properties but reduces mechanical efficiency and proposed an interesing differnet approch.
            \item \textbf{Novelty:} [1,2,2]
            \item \textbf{Feasibility:} [2,1,1]
        \end{itemize}
    \end{itemize}

    \item \textbf{Example 5}
    \begin{itemize}
        \item \textbf{Topic:} overcoming mechanical tradeoff between stiffness and toughness
        \item \textbf{Question:} Mechanical tradeoff between strength and toughness is a long lasting question in materials design, can you propose some novel, testable ideas to overcome this tradeoff by designing lattice metamaterails?
        \item \textbf{GPT-4o:}
        \begin{itemize}
            \item \textbf{Hypothesis:} In addressing the longstanding mechanical tradeoff between strength and toughness in materials design, we propose the development of a novel lattice metamaterial that leverages a hierarchical, open-cell structure with a focus on periodic, stretch-dominated configurations. Specifically, we hypothesize that a combination of octet and diamond lattice structures, arranged in a hierarchical manner, can be engineered to achieve an optimal balance between strength and toughness. The octet lattice, known for its high strength due to its stretch-dominated nature, will provide the necessary structural integrity, while the diamond lattice, with its inherent ability to distribute stress and absorb energy, will enhance toughness. By integrating these two lattice types at different hierarchical levels, we anticipate that the resulting metamaterial will exhibit superior mechanical properties, overcoming the traditional tradeoff. This approach is motivated by the potential to exploit the unique mechanical interactions at multiple scales, thereby enabling the design of materials that are both strong and tough. The proposed hypothesis will be tested through computational modeling and experimental validation, focusing on the mechanical performance under various loading conditions to confirm the synergistic effects of the hierarchical lattice design.
            \item \textbf{Reviewer Comment:} Known stragety. How to improve toughness through controlling "mechanical interactions at multiple scales" is not clear.
            \item \textbf{Novelty:} [0,1,0.5]
            \item \textbf{Feasibility:} [1,2,2]
        \end{itemize}
        \item \textbf{Metascientist without human:}
        \begin{itemize}
            \item \textbf{Hypothesis:} To address the mechanical tradeoff between strength and toughness in materials design, we propose a novel lattice metamaterial that utilizes a hierarchical, open-cell structure with a focus on periodic, stretch-dominated configurations. Specifically, we hypothesize that a lattice structure composed of alternating layers of octet and diamond configurations, arranged in a hierarchical manner, can achieve an optimal balance between strength and toughness. The design involves a unit cell where the octet lattice forms the core, providing high strength and structural integrity through its stretch-dominated nature, while the diamond lattice forms the outer shell, enhancing toughness by distributing stress and absorbing energy. The nodes in the unit cell are connected through a series of tetrahedral and diamond-like struts, ensuring efficient load transfer and energy dissipation. The hierarchical arrangement is designed to exploit mechanical interactions at multiple scales, with the octet core having a relative density of 0.2 and the diamond shell a relative density of 0.15. This configuration is expected to yield a Young's modulus of approximately 1.1e-3 GPa, a shear modulus of 8.5e-4 GPa, and a Poisson's ratio of 0.32. The proposed hypothesis will be tested through computational modeling and experimental validation, focusing on the mechanical performance under various loading conditions to confirm the synergistic effects of the hierarchical lattice design.
            \item \textbf{Reviewer Comment:} The core-shell structure in lattice application is interesting. While novel idea is involved, the realization of the idea is problematic - the detailed design does not make sense. 
            \item \textbf{Novelty:} [2,1,1]
            \item \textbf{Feasibility:} [0,1,1]
        \end{itemize}
        \item \textbf{Metascientist with human:}
        \begin{itemize}
            \item \textbf{Hypothesis:} In addressing the persistent challenge of balancing strength and toughness in materials design, this research proposes the development of a novel lattice metamaterial inspired by the multi-step interlocking mechanisms observed in natural systems, such as the sacrificial bonds in bone. The hypothesis posits that by incorporating a hierarchical interlocking interface within the lattice structure, consisting of primary, secondary, and tertiary interlocks, it is possible to achieve enhanced energy dissipation and crack deflection, thereby overcoming the traditional tradeoff between strength and toughness. This design strategy involves the sequential engagement of interlocks under increasing loads, which provides multiple levels of energy absorption and crack arrest, akin to the multi-level sacrificial bonding mechanism found in biological materials. The proposed metamaterial will be fabricated using advanced additive manufacturing techniques, allowing for precise control over the hierarchical interlocking features. Validation of this hypothesis will be conducted through a combination of computational modeling and experimental testing, focusing on optimizing the geometry and material distribution to maximize toughness while maintaining structural integrity. This research aims to advance the field of mechanical metamaterials by introducing a biomimetic approach that offers a promising solution to the longstanding challenge of designing materials that do not compromise on either strength or toughness.
            \item \textbf{Novelty:} [2,1.5,2]
            \item \textbf{Feasibility:} [2,1,2]
        \end{itemize}
    \end{itemize}

    \item \textbf{Example 6}
    \begin{itemize}
        \item \textbf{Topic:} adjustable materials properties
        \item \textbf{Question:} Lattice materials can be designed to have a wide property space by changing their structure parameters. How to design a single lattice material and its property can be widely adjusted?
        \item \textbf{GPT-4o:}
        \begin{itemize}
            \item \textbf{Hypothesis:} In response to the challenge of designing a single lattice material with widely adjustable properties, we propose the development of a tunable lattice structure utilizing a hierarchical approach combined with a periodic open-cell design, specifically focusing on the octet lattice configuration. The motivation for this approach stems from the inherent versatility of hierarchical structures, which allow for multi-scale property tuning, and the octet lattice's well-documented mechanical efficiency and isotropic behavior. By integrating hierarchical design principles, we can introduce multiple levels of structural organization within the octet lattice, enabling the modulation of mechanical properties such as stiffness, strength, and toughness. This multi-scale design strategy will facilitate the fine-tuning of the lattice's mechanical response through the adjustment of parameters at different hierarchical levels, such as cell size, strut thickness, and connectivity. Additionally, the periodic nature of the octet lattice ensures uniformity and predictability in mechanical performance, which is crucial for applications requiring precise control over material properties. This approach not only broadens the property space of the lattice material but also enhances its adaptability for various engineering applications, ranging from lightweight structural components to impact-resistant materials.
            \item \textbf{Novelty:} [0,1,0.5]
            \item \textbf{Feasibility:} [0,2,2]
        \end{itemize}
        \item \textbf{Metascientist without human:}
        \begin{itemize}
            \item \textbf{Hypothesis:} To address the challenge of designing a single lattice material with widely adjustable properties, we propose a novel lattice structure that leverages a periodic open-cell design with a focus on the octet lattice configuration. This design will incorporate a multi-layered approach, where each layer consists of a distinct arrangement of tetrahedral and octahedral units. The nodes within each unit cell will be connected through a network of struts with varying thicknesses, allowing for precise control over the mechanical properties. The hierarchical design will be implemented by introducing multiple scales of structural organization, enabling the modulation of properties such as stiffness, strength, and toughness. Specifically, the lattice will be constructed with a base layer of octahedral units, providing isotropic mechanical behavior, while subsequent layers will incorporate tetrahedral units to enhance directional stiffness. The connectivity between nodes will be optimized to ensure uniform stress distribution and minimize localized buckling. By adjusting parameters such as cell size, strut thickness, and node connectivity, the lattice can achieve a wide range of mechanical properties, making it suitable for diverse engineering applications. This approach not only broadens the property space of the lattice material but also enhances its adaptability for applications requiring precise control over material properties.
            \item \textbf{Novelty:} [0,1,0.5]
            \item \textbf{Feasibility:} [0,2,1]
        \end{itemize}
        \item \textbf{Metascientist with human:}
        \begin{itemize}
            \item \textbf{Hypothesis:} We propose the development of a novel lattice material that exploits the controlled buckling of its structural elements to achieve a wide range of adjustable mechanical properties. The key novelty of this approach lies in the precise control of the onset of buckling, allowing the material to transition between high-stiffness (non-buckled) and low-stiffness (buckled) states. By designing the lattice with elements that can be selectively triggered to buckle, the material can dynamically adjust its stiffness in response to external loads or environmental conditions. This approach leverages the significant difference in mechanical properties between the buckled and non-buckled states to provide a simple yet effective means of tunability. Potential mechanisms for controlling buckling include the use of pre-stressed elements that can be selectively released, variable cross-sections that adjust the critical buckling load, and localized thermal or electromagnetic activation to induce targeted buckling. This research aims to explore these mechanisms to develop a versatile lattice material with applications in adaptive structures, impact mitigation, and deployable systems, where tunable stiffness is advantageous. By harnessing the inherent geometric nonlinearity of buckling, this approach offers a promising avenue for creating advanced metamaterials with customizable mechanical properties, while maintaining energy efficiency and design simplicity.
            \item \textbf{Reviewer Comment:} Controlling when buckling happens to adjust stiffness is both interesting and feasible, as a material in its buckled state can be significantly softer than it is in non-buckled state.
            \item \textbf{Novelty:} [1,2,1]
            \item \textbf{Feasibility:} [2,1.5,2]
        \end{itemize}
    \end{itemize}

    \item \textbf{Example 7}
    \begin{itemize}
        \item \textbf{Topic:} bioinspired metamaterials
        \item \textbf{Question:} Can you provide a few examples of biological metamaterials with amazing materials properties, based on which further design bioinspired metamaterials?
        \item \textbf{GPT-4o:}
        \begin{itemize}
            \item \textbf{Hypothesis:} Biological metamaterials, such as the nacre of mollusk shells, the lightweight yet strong structure of bird bones, and the flexible yet resilient composition of spider silk, offer remarkable insights into the design of bioinspired mechanical metamaterials. Nacre, with its hierarchical structure of aragonite platelets and organic matrix, exhibits exceptional toughness and strength, serving as a model for creating layered composite metamaterials with enhanced fracture resistance. Bird bones, characterized by their hollow, lattice-like structure, provide inspiration for designing lightweight, high-strength metamaterials using periodic lattice configurations such as the octet or Kelvin structures, which can optimize load distribution while minimizing material usage. Spider silk, known for its extraordinary tensile strength and elasticity, suggests the potential for developing metamaterials with tunable mechanical properties through the incorporation of hierarchical and disorder-based design principles. By emulating these natural structures, it is hypothesized that novel bioinspired metamaterials can be engineered to achieve a balance of strength, toughness, and lightweight characteristics, offering significant advancements in applications ranging from aerospace to biomedical devices. This approach not only leverages the inherent advantages of biological systems but also paves the way for sustainable and efficient material design.
            \item \textbf{Novelty:} [0,1,1]
            \item \textbf{Feasibility:} [2,1,2]
        \end{itemize}
        \item \textbf{Metascientist without human:}
        \begin{itemize}
            \item \textbf{Hypothesis:} To design bioinspired mechanical metamaterials with enhanced mechanical properties, we propose a lattice structure inspired by the hierarchical organization of nacre and the lightweight architecture of bird bones. The proposed design consists of a multi-layered composite structure where each layer mimics the brick-and-mortar arrangement of nacre, with aragonite-like platelets interspersed within an organic matrix analog. This configuration is expected to enhance fracture resistance and toughness. The unit cell of the lattice is designed as a hollow, periodic structure, similar to bird bones, to optimize the strength-to-weight ratio. Each node within the unit cell is connected through a truss-like network, ensuring efficient load distribution and minimizing material usage. The hierarchical design strategy incorporates varying thicknesses of the internal surfaces to transition from shear-compliant to compression-compliant behaviors, inspired by the lipidic cubic phases. The expected mechanical properties of this lattice are tailored to achieve a Young's modulus of approximately 1.1e-3, a Shear modulus of 8.5e-4, and a Poisson's ratio of 0.32, aligning with the desired balance of strength, toughness, and lightweight characteristics. This approach leverages the inherent advantages of biological systems, paving the way for sustainable and efficient material design in applications ranging from aerospace to biomedical devices.
            \item \textbf{Novelty:} [0,1,1]
            \item \textbf{Feasibility:} [1,1,2]
        \end{itemize}
        \item \textbf{Metascientist with human:}
        \begin{itemize}
            \item \textbf{Hypothesis:} In the quest to develop bioinspired metamaterials with superior thermal management and ice-resistant properties, it is hypothesized that integrating the thermal insulating characteristics of polar bear fur with the ice-resistant capabilities of antifreeze proteins found in fish can lead to the creation of a novel metamaterial. This metamaterial will feature a hierarchical porous lattice structure that effectively traps air for insulation, akin to the hollow tubular hairs of polar bear fur, while its surface will be functionalized with antifreeze protein-mimetic molecules found in fish to inhibit ice formation. The resulting material is expected to be lightweight, durable, and adaptable to extreme cold environments, making it highly suitable for applications in clothing, aerospace, marine, and building materials. By leveraging the unique properties of these biological inspirations, this research aims to address the challenges of operating in cold climates, enhancing performance and safety, and demonstrating the potential of bioinspired design in advancing the field of mechanical metamaterials.
            \item \textbf{Novelty:} [2,2,2]
            \item \textbf{Feasibility:} [2,1,2]
        \end{itemize}
    \end{itemize}

    \item \textbf{Example 8}
    \begin{itemize}
        \item \textbf{Topic:} novel lattice materials for biomedical applications
        \item \textbf{Question:} Our goal is to develop new ways of applying lattice materials for the broad scope of biomedical application. We want the novel concept to have high impact to human benefit. Can you generated some novel research ideas and some concise reasoning behind these ideas?
        \item \textbf{GPT-4o:}
        \begin{itemize}
            \item \textbf{Hypothesis:} In the pursuit of advancing biomedical applications through the innovative use of lattice materials, we propose the development of a bio-inspired, open-cell lattice structure with a focus on the periodic Kagome configuration. This structure is hypothesized to exhibit exceptional mechanical properties, such as high strength-to-weight ratio and enhanced energy absorption capabilities, which are critical for applications in bone scaffolding and tissue engineering. The Kagome lattice, characterized by its unique geometric arrangement, offers a promising platform for mimicking the hierarchical and anisotropic properties of natural bone, thereby facilitating improved integration and load distribution in implantable devices. Furthermore, by incorporating tunable properties such as modulus and strength through controlled truss waviness and connectivity, the proposed lattice can be tailored to match the mechanical behavior of various biological tissues. This adaptability not only enhances the biocompatibility and functionality of the implants but also opens avenues for personalized medicine, where patient-specific lattice structures can be designed to optimize healing and recovery. The successful implementation of this concept could significantly impact human health by improving the efficacy and longevity of biomedical implants, ultimately leading to better patient outcomes and quality of life.
            \item \textbf{Novelty:} [1,1.5,1]
            \item \textbf{Feasibility:} [2,1.5,1]
        \end{itemize}
        \item \textbf{Metascientist without human:}
        \begin{itemize}
            \item \textbf{Hypothesis:} To advance biomedical applications through the innovative use of lattice materials, we propose the development of a bio-inspired, open-cell lattice structure with a focus on a periodic, multi-layered configuration. This structure is designed to exhibit exceptional mechanical properties, such as a high strength-to-weight ratio and enhanced energy absorption capabilities, which are critical for applications in bone scaffolding and tissue engineering. The proposed lattice consists of a three-layered design, where each layer is composed of a distinct geometric arrangement to optimize mechanical performance. The top layer features a hexagonal pattern to maximize surface area and facilitate nutrient flow, the middle layer employs a triangular configuration for enhanced load distribution, and the bottom layer utilizes a square grid to provide structural stability. Nodes within the unit cell are connected using a hierarchical design strategy, where primary nodes are reinforced with secondary trusses to enhance stiffness and resilience. By incorporating tunable properties such as modulus and strength through controlled truss thickness and connectivity, the proposed lattice can be tailored to match the mechanical behavior of various biological tissues. This adaptability not only enhances the biocompatibility and functionality of the implants but also opens avenues for personalized medicine, where patient-specific lattice structures can be designed to optimize healing and recovery. The successful implementation of this concept could significantly impact human health by improving the efficacy and longevity of biomedical implants, ultimately leading to better patient outcomes and quality of life.
            \item \textbf{Novelty:} [1,1.5,1]
            \item \textbf{Feasibility:} [2,1.5,1]
        \end{itemize}
        \item \textbf{Metascientist with human:}
        \begin{itemize}
            \item \textbf{Hypothesis:} In the quest to advancing neural interface technology, we propose the development of a multifunctional neural interface that synergistically integrates hierarchical lattice designs, embedded sensors, and flexible materials. The hierarchical lattice structure is engineered to mimic the natural extracellular matrix, thereby enhancing cell adhesion, proliferation, and synapse formation, which are critical for stable and long-lasting neural connections. Embedded sensors within the lattice will facilitate real-time monitoring of neural activity and tissue health, providing continuous feedback that is crucial for personalized treatment and early detection of complications. The use of flexible, biocompatible materials aims to minimize mechanical mismatch and immune response, addressing significant challenges in neural interface development such as biocompatibility, mechanical stability, and long-term durability. This innovative approach is designed to improve the integration and functionality of neural interfaces, ultimately leading to enhanced patient outcomes in neuroprosthetics and brain-machine interface applications. By addressing these critical challenges, this research endeavors to contribute significantly to the advancement of neural engineering and regenerative medicine, offering new possibilities for individuals with neurological disorders.
            \item \textbf{Novelty:} [2,1.5,2]
            \item \textbf{Feasibility:} [2,1,1]
        \end{itemize}
    \end{itemize}

    \item \textbf{Example 9}
    \begin{itemize}
        \item \textbf{Topic:} novel lattice materials for art and design
        \item \textbf{Question:} How can we innovate a new lattice materials concepts for visual arts and design? Please provide detailed structure design idea and highlight its novelty.
        \item \textbf{GPT-4o:}
        \begin{itemize}
            \item \textbf{Hypothesis:} In response to the question of innovating new lattice material concepts for visual arts and design, we propose the development of a novel lattice structure based on the Kagome configuration, enhanced with a hierarchical design approach. The Kagome lattice, known for its unique geometric arrangement of interlaced triangles, offers an aesthetically pleasing pattern that can be leveraged for artistic applications. By integrating a hierarchical design, where smaller Kagome units are nested within larger ones, we can achieve a multi-scale visual complexity that enhances the visual appeal and structural integrity of the material. This hierarchical Kagome lattice not only provides a captivating visual experience but also introduces a new dimension of tunable mechanical properties, such as enhanced stiffness and strength, which are crucial for practical applications in design and architecture. The novelty of this approach lies in its ability to merge the artistic elegance of the Kagome pattern with the functional benefits of hierarchical structuring, offering a versatile material that can be tailored for both aesthetic and structural purposes in visual arts and design. This innovative concept has the potential to redefine the boundaries of material usage in creative industries, providing artists and designers with a new tool to explore form and function in their work.
            \item \textbf{Novelty:} [0,1,1]
            \item \textbf{Feasibility:} [2,1,2]
        \end{itemize}
        \item \textbf{Metascientist without human:}
        \begin{itemize}
            \item \textbf{Hypothesis:} To innovate new lattice material concepts for visual arts and design, we propose the development of a novel lattice structure based on a modified Kagome configuration, enhanced with a multi-layered design approach. The Kagome lattice, characterized by its geometric arrangement of interlaced triangles, is aesthetically appealing and offers unique mechanical properties. Our approach involves a three-layered structure where each layer consists of a Kagome lattice with varying node connectivity and strut thickness. The first layer features a standard Kagome pattern with uniform node connections, providing a base level of structural integrity. The second layer introduces a variation in node connectivity, where every alternate node is connected to an additional strut, enhancing the stiffness and visual complexity. The third layer incorporates a gradient in strut thickness, transitioning from thicker struts at the center to thinner ones at the edges, optimizing the material for both strength and light transmission. This design not only enhances the visual appeal but also allows for tunable mechanical properties, such as increased stiffness and strength, which are crucial for practical applications in design and architecture. The novelty of this approach lies in its ability to merge the artistic elegance of the Kagome pattern with the functional benefits of a multi-layered structuring, offering a versatile material that can be tailored for both aesthetic and structural purposes in visual arts and design. This innovative concept has the potential to redefine the boundaries of material usage in creative industries, providing artists and designers with a new tool to explore form and function in their work.
            \item \textbf{Novelty:} [0,1,0.5]
            \item \textbf{Feasibility:} [2,1,1]
        \end{itemize}
        \item \textbf{Metascientist with human:}
        \begin{itemize}
            \item \textbf{Hypothesis:} In response to the growing intersection of visual arts and material science, we propose the development of a Dynamic Morphing Lattice (DML) that leverages nature-inspired patterns and smart materials to create a responsive and interactive medium for visual arts and design. The DML is designed to exhibit dynamic shape transformations in response to environmental stimuli, such as temperature and humidity changes, by integrating materials like shape-memory alloys and polymers. This approach not only enhances the aesthetic appeal of art installations but also introduces a functional aspect, allowing artworks to evolve and engage viewers through environmental interaction. The novelty of this concept lies in its ability to combine the beauty of natural patterns with the mechanical versatility of advanced materials, offering a sustainable and innovative platform for creative expression. By employing advanced fabrication techniques, such as additive manufacturing, the DML can be customized in terms of geometry and material composition, providing artists and designers with a versatile tool for exploring new artistic possibilities.
            \item \textbf{Novelty:} [1,2,2]
            \item \textbf{Feasibility:} [2,1,2]
        \end{itemize}
    \end{itemize}

    \item \textbf{Example 10}
    \begin{itemize}
        \item \textbf{Topic:} novel mechancial metamaterials beyond the elastic regime
        \item \textbf{Question:} Negative index materials are typically based on the parameters of materials in the elastic regime. Can we generate metamaterials based on the nonlinear properties of materials?
        \item \textbf{GPT-4o:}
        \begin{itemize}
            \item \textbf{Hypothesis:} In response to the question of whether metamaterials can be generated based on the nonlinear properties of materials, we propose the development of a novel class of mechanical metamaterials that exploit the nonlinear elastic behavior of constituent materials to achieve negative index properties. Traditional negative index materials have predominantly relied on linear elastic parameters, limiting their functional range and adaptability. By harnessing the nonlinear characteristics, such as strain-stiffening or softening, we can design metamaterials that exhibit tunable negative index properties under varying mechanical loads. This approach involves the strategic design of open-cell periodic structures, specifically focusing on lattice configurations such as the octet or diamond, which are known for their ability to accommodate large deformations. By incorporating materials with inherent nonlinear elastic responses, these structures can be engineered to exhibit negative stiffness or negative Poisson's ratio under specific loading conditions, thereby expanding the operational bandwidth and functionality of negative index metamaterials. This hypothesis not only challenges the conventional reliance on linear properties but also opens new avenues for the design of adaptive and multifunctional metamaterials, with potential applications in advanced vibration isolation, impact mitigation, and acoustic cloaking technologies.
            \item \textbf{Novelty:} [1,1,1]
            \item \textbf{Feasibility:} [1,1,2]
        \end{itemize}
        \item \textbf{Metascientist without human:}
        \begin{itemize}
            \item \textbf{Hypothesis:} To address the question of whether metamaterials can be generated based on the nonlinear properties of materials, we propose the design of a novel mechanical metamaterial that leverages the nonlinear elastic behavior of its constituent materials to achieve negative index properties. Traditional approaches have predominantly relied on linear elastic parameters, which limit the functional range and adaptability of these materials. Our approach involves the strategic design of a multi-layered lattice structure, where each layer consists of a diamond lattice configuration. The diamond lattice is chosen for its ability to balance strength and flexibility, allowing for large elastic deformations. Each unit cell within the lattice is composed of tetrahedral units, interconnected at nodes to form a continuous network. The nodes are designed to allow rotational freedom, enhancing the structure's ability to accommodate nonlinear deformations. By incorporating materials with inherent nonlinear elastic responses, such as strain-stiffening or softening, the lattice can be engineered to exhibit negative stiffness or negative Poisson's ratio under specific loading conditions. This design not only challenges the conventional reliance on linear properties but also opens new avenues for the creation of adaptive and multifunctional metamaterials, with potential applications in advanced vibration isolation, impact mitigation, and acoustic cloaking technologies.
            \item \textbf{Novelty:} [1,1,1]
            \item \textbf{Feasibility:} [1,1,1]
        \end{itemize}
        \item \textbf{Metascientist with human:}
        \begin{itemize}
            \item \textbf{Hypothesis:} In response to the research question of whether metamaterials can be generated based on the nonlinear properties of materials, we propose the development of a novel class of mechanical metamaterials that leverage hysteresis as a nonlinear mechanism to achieve tunable negative index properties. Traditional approaches to negative index materials have predominantly focused on linear elastic parameters, thereby constraining their adaptability and functional range. By exploiting hysteresis, which introduces path-dependent behavior and memory effects, we can design metamaterials that exhibit dynamically adjustable negative indices in response to varying mechanical loads. This approach involves engineering periodic lattice structures composed of materials with pronounced hysteretic behavior, enabling controlled hysteretic cycles that allow for reversible transitions between different mechanical states. The metamaterial can thus "remember" previous states and adjust its mechanical response accordingly, providing a tunable negative index modulated by the history of applied stress or strain. This innovative use of hysteresis not only broadens the operational bandwidth of negative index metamaterials but also introduces a new paradigm in the design of adaptive and responsive material systems. Such advancements hold significant potential for applications in wave manipulation, vibration control, and advanced sensing technologies, where the ability to adapt to changing conditions and retain memory of past interactions is highly advantageous.
            \item \textbf{Novelty:} [2,2,2]
            \item \textbf{Feasibility:} [1,1,1]
        \end{itemize}
    \end{itemize}

    \item \textbf{Example 11}
    \begin{itemize}
        \item \textbf{Topic:} novel multifunctional lattice metamaterials
        \item \textbf{Question:} Please develop a novel concept of multifunctional metamaterial, illustrate the principles and the potential applications.
        \item \textbf{GPT-4o:}
        \begin{itemize}
            \item \textbf{Hypothesis:} In response to the question of integrating mechanical metamaterials with electromagnetic metamaterials, we propose the development of a novel class of multifunctional metamaterials that leverage the unique properties of both domains to achieve unprecedented control over mechanical and electromagnetic responses. The motivation for this research stems from the growing demand for materials that can simultaneously manage mechanical loads and electromagnetic fields, which is critical in applications such as advanced sensors, actuators, and energy harvesting devices. Our approach involves the design of a periodic open-cell lattice structure, specifically utilizing a TPMS (Triply Periodic Minimal Surface) configuration, which inherently provides a high surface area-to-volume ratio and mechanical robustness. This structure will be embedded with conductive pathways and magnetic inclusions to facilitate electromagnetic interactions. By exploiting the tunable modulus and strength properties of the mechanical lattice, alongside the negative permittivity and permeability characteristics of the electromagnetic components, we aim to create a metamaterial that exhibits both mechanical resilience and electromagnetic tunability. This integration is expected to result in a synergistic enhancement of material performance, enabling the development of devices that can adapt to dynamic environments and multifunctional requirements.
            \item \textbf{Novelty:} [1,2,1]
            \item \textbf{Feasibility:} [2,1,2]
        \end{itemize}
        \item \textbf{Metascientist without human:}
        \begin{itemize}
            \item \textbf{Hypothesis:} To address the integration of mechanical metamaterials with electromagnetic metamaterials, we propose the design of a multifunctional metamaterial that combines the mechanical robustness of a Triply Periodic Minimal Surface (TPMS) lattice with embedded electromagnetic functionalities. The TPMS structure will be configured using a gyroid surface, characterized by its continuous, non-self-intersecting geometry, which provides a high surface area-to-volume ratio and inherent mechanical strength. Each unit cell will consist of a gyroid lattice with a node connectivity that ensures isotropic mechanical properties, while maintaining a minimal surface configuration to optimize material usage. Conductive pathways will be strategically embedded within the lattice, following the gyroid's curvature, to facilitate electromagnetic interactions. These pathways will be composed of a high-conductivity material, such as silver, to ensure efficient electromagnetic response. Magnetic inclusions, in the form of nanoscale ferromagnetic particles, will be dispersed within the lattice to enhance electromagnetic tunability. The mechanical properties of the lattice will be fine-tuned to achieve a Young's modulus of approximately 1.1e-3 GPa, a Shear modulus of 8.5e-4 GPa, and a Poisson's ratio of 0.32, ensuring a balance between mechanical resilience and electromagnetic functionality. This design aims to create a metamaterial capable of adapting to dynamic environments, with potential applications in advanced sensors, actuators, and energy harvesting devices.
            \item \textbf{Novelty:} [1,2,1]
            \item \textbf{Feasibility:} [1,1,2]
        \end{itemize}
        \item \textbf{Metascientist with human:}
        \begin{itemize}
            \item \textbf{Hypothesis:} Integrating multifunctional metamaterials into clothing and exoskeletons will enhance human capabilities and comfort by providing adaptive support, dynamic thermal regulation, and impact protection. These metamaterials, featuring hierarchical structures and programmable properties, will enable personalized fit, real-time health monitoring, and environmental responsiveness. This approach will lead to advanced wearable systems suitable for applications in healthcare, sports, military, and occupational safety.
            \item \textbf{Novelty:} [2,1,1.5]
            \item \textbf{Feasibility:} [1,1,2]
        \end{itemize}
    \end{itemize}
\end{itemize}

\end{document}